\documentclass[10pt,twocolumn,letterpaper]{article}

\usepackage[pagenumbers]{wacv}

\usepackage{multirow}

\graphicspath{{figs/}{figs/method_toy/}{figs/race5v4/bars/}{figs/race5v4/qual/}}

\newcommand{\meanstd}[2]{#1{\scriptscriptstyle \pm #2}}
\newcommand{\bmeanstd}[2]{\mathbf{#1}{\scriptscriptstyle \pm #2}}

\definecolor{wacvblue}{rgb}{0.21,0.49,0.74}
\usepackage[pagebackref,breaklinks,colorlinks,allcolors=wacvblue]{hyperref}

\def\wacvPaperID{2013}
\def\confName{WACV}
\def\confYear{2027}

\title{Multi-Axis Max@K Reinforcement Learning for\\ Representative Diversity in Text-to-Image Generation}

\author{%
Ku Onoda$^{1}$ \quad Paavo Parmas$^{1}$ \quad Hiroki Furuta \quad Soichiro Nishimori$^{1}$\\
Yuta Oshima$^{1}$ \quad Shohei Taniguchi$^{1}$ \quad Yutaka Matsuo$^{1}$\\[4pt]
$^{1}$The University of Tokyo\\
{\tt\small \{ku.onoda, paavo.parmas\}@weblab.t.u-tokyo.ac.jp}
}

\begin{document}
\maketitle
\begin{abstract}
Text-to-image (T2I) models produce realistic, prompt-aligned images, but samples for the same prompt often cover only a few visually distinct modes.
For person-centric prompts, this lack of diversity can reflect or amplify demographic skew.
We formalize representative diversity as target-mode coverage: how well a set of samples covers predefined modes, such as colors or perceived-race categories.
We then propose MultiMaxPO, a reinforcement learning post-training method that improves this coverage in diffusion T2I models by optimizing a multi-axis max@K objective.
Given one score per mode for each sample, multi-axis max@K takes the maximum over samples for each mode and then sums these maxima.
Because only the best sample counts for each mode, another high score on a mode that is already covered adds little, whereas covering a missing mode adds much more.
The objective therefore favors sets whose samples cover different modes.
We verify this effect on a synthetic Gaussian mixture and on Stable Diffusion 3.5 Medium with pixel-based color rewards, and then apply it to perceived-appearance fairness.
On held-out prompts and under three automatic evaluators, MultiMaxPO improves the Fairness Score over the base model by 0.22--0.36 on Stable Diffusion 3.5 Medium and by 0.07--0.19 on FLUX.1-dev, outperforms matched RL baselines, and maintains image quality and text alignment.
Code is available at \url{https://github.com/KuOnoda/multi-axis-maxk}.
\end{abstract}

\section{Introduction}
\label{sec:intro}

\begin{figure}[t]
  \centering
  \includegraphics[width=\linewidth]{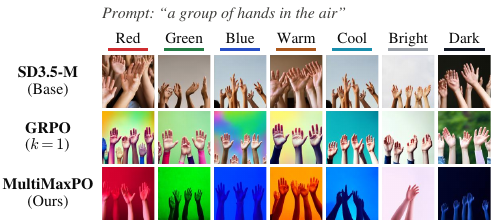}
  \caption{%
  \textbf{Diversifying color modes in T2I generation.}
  Each column shows, for one prompt, the sample that scores highest on one of seven competing pixel-based color rewards.
  MultiMaxPO covers more of the colors than the base model and GRPO ($k{=}1$).
  }
  \label{fig:rgb}
\end{figure}

Text-to-image (T2I) models can synthesize many plausible images for a prompt~\citep{rombach2022ldm,esser2024sd3}, but repeated samples for the same prompt often look alike.
Low diversity matters when users want a set of distinct options rather than a single image.
For person-centric prompts, a narrow range of appearances can also reflect or amplify the demographic skew reported in prior studies~\citep{luccioni2023stablebias,bianchi2023easily,vandewiele2025beyondprompt}.
Existing methods that increase diversity can also reduce image quality~\citep{sadat2024cads,astolfi2024cdr}.

Recent diffusion RL methods fine-tune T2I models with external image rewards~\citep{black2023ddpo,fan2023dpok,xu2023imagereward,liu2025flowgrpo,xue2025dancegrpo}.
A policy samples candidates for a prompt, a reward model scores each candidate, and the scalar rewards are used to update the policy.
Scalar rewards are well suited to objectives such as preference, image quality, and text alignment~\citep{xu2023imagereward,kirstain2023pickscore,radford2021clip}.
They are less suited to covering several mutually exclusive modes, such as object subtypes, colors, viewpoints, or perceived-appearance categories, across a set of samples.
No single image can represent all of these modes; a set is better when different samples represent different modes.

We call this goal representative diversity and formulate it as target-mode coverage. Building on max@K policy optimization~\citep{koyamada2022remax,nishimori2026emergence,tang2025optimizing,walder2025passk,bagirov2025bonworlds,takashiro2026maxk}, which trains a policy so that the best of $k$ samples scores high, we propose \emph{MultiMaxPO}, a diffusion RL method that optimizes a \emph{multi-axis max@K} objective.
For each prompt, a fixed reward function scores every generated image on each target mode.
The objective scores a set of $k$ samples by taking the best score in the set for each mode and then summing over modes.
Unlike rewards that first sum the scores within each image, this order makes another high score on an already covered mode worth little, so samples gain more by covering different modes.
Figure~\ref{fig:rgb} shows the effect on seven competing color rewards: after training, a sampled batch contains a strong example of each color.

To optimize this objective, we apply the max@K advantage estimator of MaxPO~\citep{takashiro2026maxk} separately to each mode, sum the per-mode advantages, and use them in a GRPO-style diffusion update~\citep{black2023ddpo,liu2025flowgrpo,xue2025dancegrpo}.

We evaluate the method in three settings.
First, in a synthetic Gaussian mixture, where modes and rewards are known, MultiMaxPO moves probability mass toward all target modes, while a scalar reward for one mode concentrates mass on that mode.
Second, we train Stable Diffusion 3.5 Medium (SD3.5-M)~\citep{esser2024sd3} with simple color and brightness rewards computed directly from pixels, testing whether the method separates competing reward axes when no learned model decides the modes.
Third, we apply the method to perceived-appearance fairness on SD3.5-M and FLUX.1-dev, with perceived-race, perceived-gender, and joint categories as target modes.

On held-out prompts, MultiMaxPO improves the Fairness Score for perceived race and race-by-gender cells over the base model and matched RL baselines under three automatic evaluators~\citep{radford2021clip,wu2025rewarddance,karkkainen2021fairface}, two of which are not used in training, while image-quality~\citep{kirstain2023pickscore,schuhmann2022laion} and text-alignment~\citep{radford2021clip} metrics are maintained.
Our main contributions are as follows.
\begin{itemize}\itemsep2pt \parskip0pt \topsep2pt
\item We formulate representative diversity in T2I generation as target-mode coverage and introduce multi-axis max@K, a special case of our earlier ReMax objective~\citep{koyamada2022remax,nishimori2026emergence} applied to diversity in generative models (Sec.~\ref{sec:related}).
\item We turn this objective into a diffusion RL algorithm, MultiMaxPO, which applies the MaxPO estimator~\citep{takashiro2026maxk} per axis and keeps the modes separate until the final sum.
\item We validate the method on a synthetic Gaussian mixture, pixel-based color rewards, and perceived-appearance fairness on two T2I models.
\end{itemize}

\section{Related Work}
\label{sec:related}

\paragraph{RL post-training for T2I.}
RL post-training adapts T2I models to external scalar rewards, such as learned preference models~\citep{xu2023imagereward}.
DDPO and DPOK cast denoising as a sequential decision process and apply policy
gradients~\citep{black2023ddpo,fan2023dpok}, preference optimization has been
adapted to diffusion~\citep{wallace2024diffusiondpo}, and Flow-GRPO and its
successors bring group-relative updates to flow-based
generators~\citep{liu2025flowgrpo,xue2025dancegrpo,he2025tempflowgrpo}.
We build on the Flow-GRPO training setup.

\paragraph{Set-level, best-of-$K$, and vector rewards.}
ReMax~\citep{koyamada2022remax,nishimori2026emergence} evaluates a policy by the expected maximum return over $K$ retries, with the
reward function drawn from a distribution, and shows that optimizing this
set-level objective leads to emergent exploration.
These works also provide a policy gradient estimator that extends to arbitrary
objectives over sets or trajectory trees.
Our multi-axis max@K objective is the special case in which the reward function
is one of the reward axes, drawn at random.
We apply this objective to diversity in image generation, whereas ReMax studied
exploration in simpler RL tasks.
Follow-ups extend ReMax to continuous actions~\citep{nishimori2026retry} and to a
bandit analysis proving sublinear regret~\citep{tong2026finite}. \citet{gx2026using} later proposed a closely related objective with more focus
on generative modeling.
In LLM post-training, several works directly optimize pass@K, the special case
of max@K with binary rewards~\citep{tang2025optimizing,walder2025passk,chow2024bonft,chen2025passktraining}, and \citet{bagirov2025bonworlds} optimize max@K itself.
We use the max@K gradient estimator of MaxPO~\citep{takashiro2026maxk}.
Set-level objectives have also been used to preserve answer
diversity~\citep{li2026setpo,puri2026escaping,hamid2026polychromic}, and
\citet{parmas2026ordergrad} generalize the max aggregator to arbitrary order
statistics.
Concurrent VPO~\citep{bahlousboldi2026vpo} uses vector-valued rewards so that
samples specialize across reward trade-offs in LLM tasks; its objective can also
be viewed as a special case of ReMax.
We instead target coverage of a fixed set of modes in T2I generation and take
the maximum separately on each reward axis before summing over axes.

\paragraph{Diversity, fairness, and inclusion in T2I generation.}
Coverage and recall measures for generative models
\citep{sajjadi2018assessing,kynkaanniemi2019improved,naeem2020reliable}
and diffusion diversity rewards
\citep{miao2024diversity,liu2026diversegrpo,liu2026drift}
measure or improve overall diversity; we instead target coverage of predefined
target modes.
T2I fairness work includes bias evaluations and benchmarks
\citep{luccioni2023stablebias,bianchi2023easily,
vandewiele2025beyondprompt,dinca2024openbias,
lee2023heim,chinchure2024tibet,karkkainen2021fairface},
prompt- and embedding-based methods
\citep{zhang2023itigen,fu2025fairimagen},
guidance and model-internal editing or steering methods
\citep{friedrich2023fairdiffusion,brack2023sega,
gandikota2024uce,kim2025weak,parihar2024balancing,
li2024interpretdiffusion,sreelatha2025respodiff},
and training-time fairness objectives
\citep{shen2024finetuning,li2025fairmapping,jiang2024fairgen,huang2025ddm}.
RAIGen~\citep{sreelatha2026raigen} discovers rare attributes from model-internal
representations without predefined categories.
Our method assumes predefined target modes and addresses how to assign
policy-gradient credit within a group of samples from per-mode scores. The closest RL method is concurrent Fair-GRPO~\citep{chen2026holofair}, which
assigns credit by counting hard category labels; our GRPO (count) baseline follows this design.

\section{Diversity as a Multi-Axis Set Objective}
\label{sec:formulation}

\paragraph{Preliminaries.}
A text-to-image generator induces a prompt-conditional distribution
$\pi_\theta(x\mid c)$ over images $x$ for each prompt $c$. For a fixed prompt,
let $S_k(c)=\{x_1,\ldots,x_k\}$ denote $k$ independent samples from
$\pi_\theta(\cdot\mid c)$. We assume $D$ predefined target modes for each prompt, indexed by $d=1,\ldots,D$.

\subsection{Diversity as mode coverage}
Representative diversity is a property of a set of samples, not of a single image: the question is whether the generator produces at least one clear example of each target mode.
For a prompt $c$, the generator should give each mode enough probability that it appears in a finite sample set with high probability.
Because one image usually represents at most one of several mutually exclusive modes, we define the objective on a sample set drawn from $\pi_\theta(\cdot\mid c)$ and ask which target modes it contains.

\subsection{Mode coverage as multi-axis max@K}
To turn target-mode coverage into a training objective, we first consider the
hard-label case. Suppose that each image belongs to exactly one target mode
$g(x,c)\in\{1,\dots,D\}$. Mode $d$ is covered by $S_k(c)$ if at least one sample
in it belongs to mode $d$. The indicator
of this event is
\begin{equation}
\begin{aligned}
&\mathbf{1}\{\exists x_i\in S_k(c): g(x_i,c)=d\}\\
&\quad= \max_{x_i\in S_k(c)} \mathbf{1}\{g(x_i,c)=d\}.
\end{aligned}
\label{eq:coverage-event}
\end{equation}
In the hard-label setting, the coverage score of the set is then the weighted
number of target modes that appear in it:
\begin{equation}
H_k(S_k,c) = \sum_{d=1}^{D} w_d(c) \max_{x_i\in S_k(c)} \mathbf{1}\{g(x_i,c)=d\}.
\label{eq:hard-coverage-score}
\end{equation}
The weights $w_d(c)$ set the relative importance of the modes; in our
experiments, they are uniform.

In practice, mode membership is often measured by soft scores rather than hard
labels. For each image $x$ and prompt $c$, let $r_d(x,c)\in[0,1]$ denote the
score for target mode $d$, computed by a fixed reward function, such as a pixel-based rule or a classifier. We define
multi-axis max@K as the soft-score extension of the hard coverage score:
\begin{equation}
G_k(S_k,c)=\sum_{d=1}^{D} w_d(c)\max_{x_i\in S_k(c)} r_d(x_i,c).
\label{eq:multi-axis-set-score}
\end{equation}
For each mode, the maximum picks the best sample in the set, so the set scores high on mode $d$ when at least one sample has a high $r_d$, without assigning a hard category.

The key step is taking the maximum on each axis before summing over axes.
A scalar-first score
$G_{\mathrm{scalar}}(S_k,c)=\max_{x_i\in S_k}\sum_d w_d(c)r_d(x_i,c)$ collapses the axes before samples are compared.
In contrast, Eq.~\eqref{eq:multi-axis-set-score} allows samples to represent different modes. This order matters when high scores on different axes compete, as in mutually exclusive mode labels; when the axis scores sum to one and the weights are uniform, the scalar-first score is constant.

With hard rewards, $G_k$ coincides with $H_k$, and its expectation admits a
closed form.

\noindent\textbf{Proposition 1 (target-mode coverage and optimum).}
For mutually exclusive hard mode rewards $r_d(x,c)=\mathbf{1}\{g(x,c)=d\}$ and independent samples from $\pi_\theta(\cdot\mid c)$ with mode probabilities $q_d(c)=P(g(x,c)=d)$,
\begin{equation}
\mathbb{E}[G_k(S_k,c)]=
\sum_d w_d(c)\left[1-\left(1-q_d(c)\right)^k\right].
\label{eq:hard-mode-coverage}
\end{equation}
With uniform positive weights and $k>1$, this objective is strictly concave over the probability simplex and is uniquely maximized by $q_d(c)=1/D$ for all $d$. With uniform weights and $k{=}1$, it is constant.

The term $1-(1-q_d(c))^k$ is the probability that mode $d$ appears at least once in a $k$-sample set, so maximizing Eq.~\eqref{eq:multi-axis-set-score} directly increases expected mode coverage. Its marginal gain, $k(1-q_d)^{k-1}$, is larger for lower-probability modes when $k>1$. This pushes mass toward underrepresented modes and yields the uniform optimum in the unregularized, mutually exclusive hard-label case.
Since the objective is constant for $k{=}1$ (Proposition~1), $k>1$ is what makes it reward coverage.

\subsection{Pooled composition and finite-batch coverage}
\label{sec:relation}

A task may also specify a target composition pooled over samples.
Existing debiasing methods often align the expected mode distribution of generated
images, computed from hard labels or soft assignment scores, with a target distribution
\citep{shen2024finetuning,zhang2023itigen,fu2025fairimagen}.
When the target is uniform over the $D$ target modes, we summarize the pooled balance with the Fairness Score~\citep{fu2025fairimagen}, defined as
\begin{equation}
F=1-\frac{\sum_d|p_d-1/D|}{2(1-1/D)}\in[0,1],
\label{eq:fairness}
\end{equation}
where $p_d$ is the mode frequency pooled over the evaluated prompts and $F{=}1$ means the frequencies are uniform.

Pooled composition and finite-batch coverage are related but not identical.
Under independent hard-label sampling with $q_d(c)$ as in Proposition~1, mode $d$
appears at least once in an $M$-sample evaluation batch with
probability $1-(1-q_d(c))^M$, whose derivative $M(1-q_d)^{M-1}$ is largest
for low-probability modes.
Thus, a small increase in the probability of a rare mode can substantially raise its chance of appearing in a finite batch. We report both pooled composition and finite-batch coverage (Sec.~\ref{sec:experiments}, Appendix~\ref{sec:app-finite-visibility}).

\begin{figure*}[t]
  \centering
  \includegraphics[width=\textwidth]{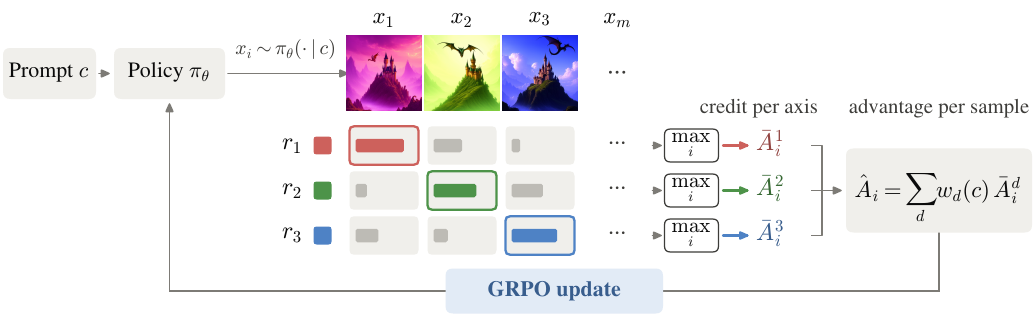}
  \caption{\textbf{MultiMaxPO.} For each prompt, the policy generates $m$ candidates, and each is scored on $D$ reward axes $r_d$, one per target mode. In this example, the axes score red, green, and blue; in the fairness experiments, they score perceived-race and perceived-gender categories. Multi-axis max@K compares samples separately on each axis before summing over axes, so different samples receive credit for different modes: $x_1$, $x_2$, and $x_3$ each have the best score on one axis (outlined). The per-axis credits $\bar A_i^d$ are summed into per-sample advantages $\hat A_i$ for a GRPO-style update. The maximum over samples is used only in training; at evaluation, images are sampled from the fine-tuned model as usual.}
  \label{fig:method}
\end{figure*}

\section{Training with MultiMaxPO}
\label{sec:method}

\subsection{From set coverage to an RL objective}
\label{sec:objective}

Section~\ref{sec:formulation} formalizes representative diversity as the coverage score of a sampled set,
$G_k(S_k,c)=\sum_d w_d(c)\max_{x_i\in S_k}r_d(x_i,c)$. We turn this score into an RL objective for a diffusion or flow model (Fig.~\ref{fig:method}): for each prompt, the model generates $m$ candidates, each candidate is scored on the $D$ axes, and the scores are converted into per-sample advantages one axis at a time.

The training objective is
\begin{equation}
\begin{aligned}
J(\theta)
&=\mathbb{E}_c\,\mathbb{E}_{S_k\sim\pi_\theta(\cdot\mid c)^k}
\big[G_k(S_k,c)\big]\\
&\quad-\beta\,\mathbb{E}_c
\big[\mathrm{KL}(\pi_\theta(\cdot\mid c)\Vert\pi_0(\cdot\mid c))\big],
\end{aligned}
\label{eq:maxk}
\end{equation}
where $\pi_0$ is the base generator and $\beta > 0$ is the weight of the Kullback--Leibler (KL) penalty, as in Flow-GRPO~\citep{liu2025flowgrpo}. In a diffusion or flow model, each image $x_i$ is the endpoint of a generated trajectory, and the per-sample advantage computed below is applied to that trajectory's policy gradient. Appendix~\ref{sec:app-diffrl} reviews this diffusion RL setup.

We use three sample sizes. The objective is defined for sets of size $k$; each training prompt generates $m\geq k$ candidates to estimate per-sample credit; and evaluation uses $M$ samples per prompt. Only $k$ and $m$ affect training.

\subsection{Per-axis credit from max@K}
\label{sec:credit}

The simplest case is $m{=}k\ge2$. Let
$v_d(S_k,c)=\max_{x_i\in S_k} r_d(x_i,c)$ be the best score on axis $d$. The leave-one-out marginal contribution of sample $x_i$ on axis $d$ is
\begin{equation}
\begin{aligned}
A_i^d
&= v_d(S_k,c)-v_d(S_k\!\setminus\!\{x_i\},c)\\
&= \Big[r_d(x_i,c)-\max_{j\ne i}r_d(x_j,c)\Big]_+ .
\end{aligned}
\label{eq:axis-credit}
\end{equation}
A sample receives nonzero credit on axis $d$ only when removing it decreases the maximum on that axis. In Fig.~\ref{fig:method}, for example, only $x_1$ receives credit on $r_1$, equal to its margin over the runner-up $x_2$; $x_2$ and $x_3$ receive credit on $r_2$ and $r_3$ in the same way.

Note that $G_k(S_k,c)=\sum_d w_d(c)v_d(S_k,c)$. Define the baseline
$b_i=\sum_d w_d(c)v_d(S_k\!\setminus\!\{x_i\},c)$. Since $b_i$ does not depend on $x_i$,
subtracting it keeps the gradient unbiased:
\begin{equation}
\begin{aligned}
&\nabla_\theta\,\mathbb{E}\big[G_k(S_k,c)\big]\\
&\quad=\mathbb{E}\Big[
\sum_i \nabla_\theta\log\pi_\theta(x_i\mid c)\,
\big(G_k(S_k,c)-b_i\big)
\Big].
\end{aligned}
\label{eq:unbiased}
\end{equation}
The term $G_k(S_k,c)-b_i=\sum_d w_d(c)A_i^d$ is the weighted sum of the per-axis credits.

\begin{figure*}[t]
  \centering
  \includegraphics[width=0.9\textwidth]{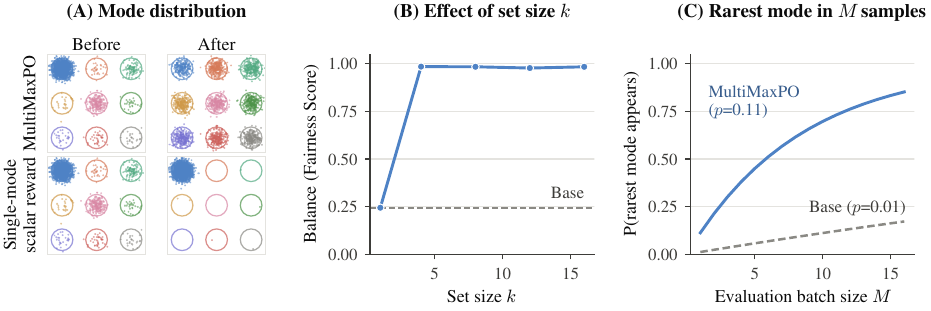}
  \caption{\textbf{Synthetic experiment with nine Gaussian modes.}
  \textbf{(A)} Mode distribution before and after training ($k{=}16$). From the same skewed base distribution, a scalar reward for one mode concentrates probability mass on that mode, whereas MultiMaxPO
  spreads mass across all nine modes.
  \textbf{(B)} Fairness Score after training for each set size $k$:
  it stays at the base level for $k{=}1$ and is close to 1 for $k\geq4$.
  \textbf{(C)} Probability $1-(1-p)^M$ that the rarest mode appears at least once among $M$ samples; $p{=}0.01$ before and $0.11$ after training.}
  \label{fig:toy}
\end{figure*}

\subsection{Practical EI+L2O estimator}
\label{sec:practical-estimator}

In training, each prompt produces an $m$-sample candidate group $S_m$, and the
main experiments use $1<k<m$. Per-sample credit is then estimated by averaging improvements over $k$-sample subsets of the candidates; for this, we apply the expected-improvement (EI) estimator with the leave-two-out (L2O) baseline of MaxPO~\citep{takashiro2026maxk} on each axis, hence the name MultiMaxPO.

For axis $d$, EI compares candidate $x_i$ against
size-$(k{-}1)$ comparison subsets $T$ drawn from the other candidates:
\begin{equation}
s_i^d(k)=
\mathbb{E}_{T}
\Big[\big(r_d(x_i,c)-\max_{x_j\in T}r_d(x_j,c)\big)_+\Big],
\label{eq:raw-ei}
\end{equation}
where the expectation is computed over all such subsets.

The L2O baseline averages the EI scores of the other samples
while excluding $x_i$ from their comparison subsets:
\begin{equation}
b_i^d(k)=\frac{1}{m-1}\sum_{j\ne i}s_{j,-i}^d(k),
\label{eq:l2o-baseline}
\end{equation}
where $s_{j,-i}^d(k)$ is Eq.~\eqref{eq:raw-ei} for sample $j$ with
comparison subsets drawn from $S_m\setminus\{x_i,x_j\}$. This yields the raw per-axis advantage $\tilde A_i^d=s_i^d(k)-b_i^d(k)$. L2O requires $k\leq m-1$.
Unlike Eq.~\eqref{eq:axis-credit}, this advantage can be negative: a sample whose expected improvement falls below its baseline, such as one that repeats a mode another sample already covers, is pushed down.

For optimization, we divide the raw per-axis advantages by their standard deviation within each prompt group:
\begin{equation}
\bar A_i^d=\operatorname{sg}\!\left(
\frac{\tilde A_i^d}{\operatorname{std}_{j}(\tilde A_j^d)+\epsilon}
\right).
\label{eq:centered-ei}
\end{equation}
Here $\operatorname{sg}(\cdot)$ denotes the stop-gradient operator. The raw EI+L2O advantages already have zero mean within the group under exact subset averaging, so we only rescale them, which puts the axes on a common scale before they are summed. The per-sample advantage is then
\begin{equation}
\hat A_i=\sum_d w_d(c)\,\bar A_i^d.
\label{eq:axis-aggregate}
\end{equation}

For the $k{=}1$ baseline, each axis uses the image reward standardized within the prompt group, and the per-axis advantages are summed; this is GRPO with each image scored on its own rather than as part of a set.

\section{Controlled Experiments}
\label{sec:diagnostics}
\subsection{Synthetic 2-D Gaussian modes}
\label{sec:toy}
Before the T2I experiments, we test the credit-assignment mechanism without image generation or learned classifiers.
A categorical policy over nine two-dimensional Gaussian modes starts from a skewed distribution; each mode is a separate reward axis, and we test whether MultiMaxPO moves probability mass toward all nine.

As a baseline, we use a scalar reward for one preferred mode, $R_{\mathrm{scalar}}(g)=\mathbf{1}\{g{=}1\}$, where $g$ is the sampled mode index; summing the one-hot rewards over all modes would give every sample the same reward. Starting from the same skewed base policy, the scalar baseline concentrates on its preferred mode, whereas MultiMaxPO spreads mass across all nine modes (Fig.~\ref{fig:toy}A). The Fairness Score stays at the base level for $k{=}1$ and is close to 1 for $k\geq4$ (Fig.~\ref{fig:toy}B), and the probability of the rarest mode rises from 0.01 to 0.11, so it is far more likely to appear in a finite batch (Fig.~\ref{fig:toy}C).

\subsection{Rule-based color modes in T2I}
\label{sec:pixel-bridge}
Next, we train SD3.5-M with seven color rewards computed directly from image pixels: red, green, blue, warm, cool, bright achromatic, and dark achromatic.
These rewards form a soft partition of image pixels: each pixel is softly assigned among the competing axes, and the image-level reward for an axis is the average assignment over pixels.
Because the assignments sum to one, a single image cannot score high on every color.
The rewards follow a fixed rule, so no learned model decides the colors.
During training, each color score is multiplied by PickScore~\citep{kirstain2023pickscore} to discourage low-quality images; Appendix~\ref{sec:app-diagnostics} gives the details.

Evaluation uses held-out Pick-a-Pic prompts~\citep{kirstain2023pickscore}.
For each evaluation prompt $c\in\mathcal{P}$, we draw $M{=}12$ samples and average the best score in the batch for each color over colors and prompts:
\begin{equation}
C_{\mathrm{pix}}=
\frac{1}{|\mathcal{P}|D}
\sum_{c\in\mathcal{P}}\sum_{d=1}^{D}
\max_{x_i\in S_M(c)} r_d(x_i,c),
\label{eq:pixel-batchmax}
\end{equation}
with $D{=}7$. $C_{\mathrm{pix}}$ is $0.236$ for Base, $0.227$ for GRPO ($k{=}1$), and $0.370$ for MultiMaxPO ($k{=}7$), a $57\%$ increase over Base; GRPO ($k{=}1$) does not improve it.
Figure~\ref{fig:rgb} shows the best sample per color for one prompt.

\begin{table*}[t]
  \centering
  \caption{\textbf{Results on SD3.5-M.} Fairness Score $F$ on 200 held-out occupation prompts under the CLIP (training reward), VLM, and FairFace (FF) evaluators (higher is closer to uniform), with image-quality and text-alignment metrics, for race5 and the ten race5$\times$gender2 cells. RL rows report mean $\pm$ standard deviation over three runs. Pick., Aes., and C-T denote PickScore, aesthetic score, and CLIP-T.}
  \label{tab:main-fairness}
  \footnotesize
  \renewcommand{\arraystretch}{1.12}
  \setlength{\tabcolsep}{2.7pt}
  \begin{tabular}{@{}lcccccc@{\hspace{10pt}}cccccc@{}}
    \toprule
    & \multicolumn{6}{c}{race5} & \multicolumn{6}{c}{race5$\times$gender2}\\
    \cmidrule(r{8pt}){2-7}\cmidrule{8-13}
    & \multicolumn{3}{c}{Fairness $F$} & \multicolumn{3}{c}{Quality} & \multicolumn{3}{c}{Fairness $F$} & \multicolumn{3}{c}{Quality}\\
    \cmidrule(lr){2-4}\cmidrule(r{8pt}){5-7}\cmidrule(lr){8-10}\cmidrule(l){11-13}
    Method & CLIP & VLM & FF & Pick. & Aes. & C-T & CLIP & VLM & FF & Pick. & Aes. & C-T\\
    \midrule
    SD3.5-M (Base) & 0.423 & 0.286 & 0.494 & 0.784 & 5.42 & 0.222 & 0.487 & 0.424 & 0.530 & 0.784 & 5.42 & 0.222\\
    FairImagen~\citep{fu2025fairimagen} & 0.446 & 0.320 & 0.516 & 0.731 & 5.42 & 0.152 & 0.517 & 0.478 & 0.566 & 0.731 & 5.42 & 0.152\\
    Weak Guidance~\citep{kim2025weak} & 0.613 & 0.612 & 0.738 & 0.733 & 5.51 & 0.156 & 0.653 & 0.693 & 0.707 & 0.734 & 5.53 & 0.157\\
    GRPO ($k{=}1$)~\citep{liu2025flowgrpo} & $\meanstd{0.450}{0.008}$ & $\meanstd{0.285}{0.007}$ & $\meanstd{0.521}{0.007}$ & 0.769 & 5.28 & 0.218 & $\meanstd{0.546}{0.010}$ & $\meanstd{0.486}{0.006}$ & $\meanstd{0.557}{0.011}$ & 0.761 & 5.16 & 0.215\\
    GRPO (count)~\citep{chen2026holofair} & $\meanstd{0.479}{0.008}$ & $\meanstd{0.385}{0.012}$ & $\meanstd{0.561}{0.013}$ & 0.784 & 5.43 & 0.222 & $\meanstd{0.539}{0.010}$ & $\meanstd{0.542}{0.019}$ & $\meanstd{0.592}{0.016}$ & 0.786 & 5.46 & 0.223\\
    \textbf{Ours} & $\bmeanstd{0.659}{0.016}$ & $\bmeanstd{0.644}{0.015}$ & $\bmeanstd{0.771}{0.013}$ & \textbf{0.793} & \textbf{5.56} & \textbf{0.225} & $\bmeanstd{0.718}{0.024}$ & $\bmeanstd{0.781}{0.007}$ & $\bmeanstd{0.746}{0.009}$ & \textbf{0.794} & \textbf{5.56} & \textbf{0.224}\\
    \bottomrule
  \end{tabular}
\end{table*}

\section{MultiMaxPO for Perceived-Appearance Fairness}
\label{sec:experiments}

Perceived-appearance fairness concerns skews in visible attributes, such as perceived race and gender, in images generated from person-centric prompts~\citep{chuang2023debias, kim2025weak, jiang2024fairgen, fu2025fairimagen, li2025fairmapping, chen2026holofair}.
Fairness is a property of the output distribution, not of a single image, so we treat these categories as target modes.

\subsection{Experimental setup}
\label{sec:setup}

\paragraph{Problem setting.}
Following prior work~\citep{kim2025weak,chuang2023debias}, we evaluate three label spaces: five-category perceived race $\{\text{White}, \text{Black}, \text{Asian}, \text{Indian}, \text{Latino}\}$ (race5), binary perceived gender (gender2), and their joint cells (race5$\times$gender2).
Asian denotes East/Southeast Asian, and Indian denotes South Asian.
These categories follow prior automatic evaluation protocols and are not a complete taxonomy of race or ethnicity; other presentations, such as Middle Eastern, Indigenous, or multiracial, could be added as reward axes given a reliable score.
All labels are automatic estimates of perceived attributes (Appendix~\ref{sec:app-ethics}).

\paragraph{Training.}
Rollouts use only 10 denoising steps, so face detectors may find no face in rollout images. We therefore score every rollout image with a frozen CLIP prompt ensemble: a softmax over the image's similarity to category text prototypes gives soft perceived-race scores, and perceived-gender scores use one prompt per category. The race5$\times$gender2 reward is the outer product of the two (Appendix~\ref{sec:app-probes}).
All RL methods fine-tune SD3.5-M with LoRA on 1{,}080 neutral-person training prompts, with KL $\beta{=}0.05$ and $k{=}D$ unless otherwise stated, and results are averaged over three training runs.

\paragraph{Baselines.}
All RL methods share the backbone, prompts, rollout budget, and reward model, and differ only in the credit rule. GRPO ($k{=}1$) (Sec.~\ref{sec:practical-estimator}) scores each image on its own, and GRPO (count) uses Fair-GRPO-style hard-assignment count credit~\citep{chen2026holofair}, counting the joint cells on race5$\times$gender2. We also evaluate FairImagen~\citep{fu2025fairimagen} and Weak Guidance~\citep{kim2025weak} with the settings that give the highest Fairness Score in our sweeps (Appendix~\ref{sec:app-config}).

\paragraph{Evaluation.}
Evaluation uses 200 held-out occupation prompts with $M{=}16$ generated images per prompt.
The main metric is the pooled Fairness Score in Eq.~\eqref{eq:fairness}.

We report results using three automatic evaluators: the CLIP prompt ensemble~\citep{radford2021clip} used during training, a yes/no evaluator using the vision-language model (VLM) Qwen2.5-VL~\citep{bai2025qwen25vl,wu2025rewarddance}, and an independent FairFace classifier~\citep{karkkainen2021fairface}. The VLM and FairFace evaluators are not used for training and therefore test whether gains transfer beyond the CLIP reward.
Appendix~\ref{sec:app-probes} gives the scoring rules, prompts, face detection, and label mapping.

To check image quality and text alignment, we also report PickScore~\citep{kirstain2023pickscore}, aesthetic score~\citep{schuhmann2022laion}, and CLIP-T~\citep{radford2021clip}.

\subsection{Results}
\label{sec:main}

\begin{figure*}[t]
  \centering
  \includegraphics[width=\textwidth]{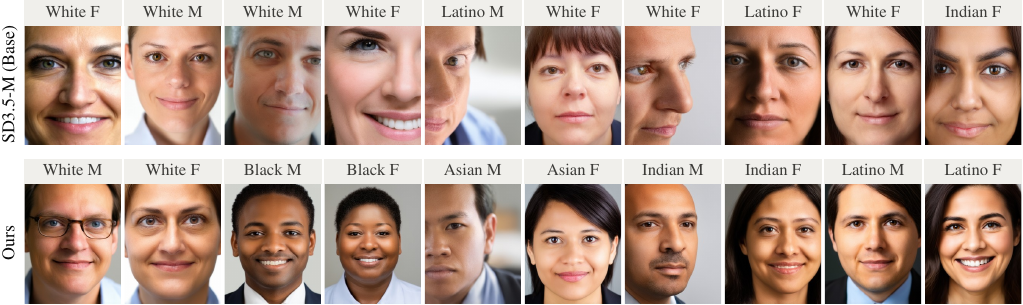}
  \caption{\textbf{Held-out examples on race5$\times$gender2.} For the \emph{underwriter} prompt, one Ours image for each joint cell and the Base image from the same seed, with automatic FairFace labels. In the full 16-image batch, Base covers 5 of the 10 cells and Ours covers all 10.}
  \label{fig:race5-qual-batch}
\end{figure*}

\paragraph{MultiMaxPO outperforms the matched RL baselines.}
Table~\ref{tab:main-fairness} (left) compares methods on race5.
Ours improves the Fairness Score over the strongest matched baseline, GRPO (count),
by $+0.18$ (CLIP), $+0.26$ (VLM), and $+0.21$ (FairFace), with 95\% prompt-bootstrap intervals above zero (Appendix~\ref{sec:app-learning-curve}), while GRPO ($k{=}1$) stays near the base level.
Ours also scores above Weak Guidance, the strongest external baseline, under all three evaluators.

\begin{figure}[t]
  \centering
  \includegraphics[width=\linewidth]{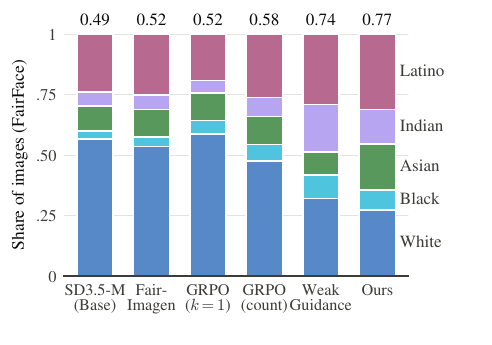}
  \caption{\textbf{Perceived-race composition on held-out prompts.} FairFace~\citep{karkkainen2021fairface} top-label shares on race5 for one run of each method, sorted by Fairness Score; labels above the bars give the FairFace Fairness Score. Images without a detected face are excluded; Appendix~\ref{sec:app-composition} lists the shares.}
  \label{fig:race5-distribution-bars}
\end{figure}

\paragraph{Gains hold under evaluators not used for training.}
Under the two evaluators not used for training, the Fairness Score also rises: VLM from $0.286$ to $0.644$ and FairFace from $0.494$ to $0.771$.
Figure~\ref{fig:race5-distribution-bars} shows the FairFace
composition for one run of each method: Base and GRPO ($k{=}1$) remain dominated by
the White label, GRPO (count) and FairImagen move only modestly, and
Ours is closest to uniform.

\paragraph{MultiMaxPO keeps image quality and text alignment.}
FairImagen and Weak Guidance improve pooled fairness but lower PickScore
($0.784\to0.731/0.733$) and text alignment (CLIP-T $0.222\to0.152/0.156$). Ours keeps PickScore,
aesthetic score, and CLIP-T at or slightly above the base model.
On 80 prompts without people, Ours does not add people to images (Appendix~\ref{sec:app-quality}).

\paragraph{Coverage extends to ten joint cells.}
On race5$\times$gender2, Ours scores highest under all three evaluators, above both GRPO (count) and Weak Guidance (Table~\ref{tab:main-fairness}, right). This setting matters because a balanced gender marginal can coexist with a skewed joint distribution; on gender2 alone, the base model is already close to uniform (Appendix~\ref{sec:app-learning-curve}).
Finite-batch coverage improves less than pooled balance (Appendix~\ref{sec:app-finite-visibility}). On race5, all five categories appear in 19\% of Ours' 16-image batches (Base 9\%, uniform generator 86\%). On race5$\times$gender2, the rarest cell appears in 23.8\% of batches (Base 13.5\%, GRPO (count) 16.8\%); all ten cells rarely appear together, even for a uniform generator (7\%).

\paragraph{Qualitative view.}
Figure~\ref{fig:race5-qual-batch} shows images from the race5$\times$gender2 models for the held-out prompt \emph{underwriter}; the images are selected by the fixed rule in Appendix~\ref{sec:app-qual-selection}.
In the 16-image batch, 11 of the 16 Base images are labeled White, while the Ours images spread over all five race5 categories and both genders.
Different samples in the same batch cover different target modes, which is what MultiMaxPO trains for.
Appendix~\ref{sec:app-qual-selection} shows complete race5 batches for six held-out prompts.

\paragraph{Why count credit falls short.}
GRPO (count) improves only modestly over the base model, which points to two limitations of hard-assignment credit: top-label assignment discards partial evidence for rare categories, and a category to which no sample is assigned receives no credit.
MultiMaxPO instead keeps the soft score of each category and compares samples on each axis, so a category to which no sample is assigned still gives credit to the samples that score highest on it. Softening
the count reward does not close the gap (Appendix~\ref{sec:app-learning-curve}).

\paragraph{Results on FLUX.1-dev.}
We repeat the race5 and race5$\times$gender2 experiments on FLUX.1-dev~\citep{blackforestlabs2024flux} with the same training prompts, reward, and evaluators (Appendix~\ref{sec:app-flux}).
In both settings, Ours scores higher than the base model and GRPO (count) under all three evaluators (Table~\ref{tab:flux-main}), and image quality stays about the same (PickScore $0.805$ for Base and $0.805$/$0.803$ for Ours).
The gains are smaller than on SD3.5-M, where each training step uses eight times as many images.

\begin{table}[t]
  \centering
  \caption{\textbf{Results on FLUX.1-dev.} Fairness Score $F$; training prompts, reward, and held-out prompts as in Table~\ref{tab:main-fairness}. RL rows report mean $\pm$ std over three runs.}
  \label{tab:flux-main}
  \footnotesize
  \renewcommand{\arraystretch}{1.12}
  \setlength{\tabcolsep}{5pt}
  \begin{tabular}{@{}lccc@{}}
    \toprule
    Method & CLIP & VLM & FF\\
    \midrule
    \multicolumn{4}{@{}l}{\emph{race5}}\\
    FLUX.1-dev (Base) & 0.331 & 0.117 & 0.387\\
    GRPO ($k{=}1$) & $\meanstd{0.299}{0.010}$ & $\meanstd{0.082}{0.012}$ & $\meanstd{0.364}{0.030}$\\
    GRPO (count) & $\meanstd{0.397}{0.022}$ & $\meanstd{0.198}{0.020}$ & $\meanstd{0.474}{0.011}$\\
    \textbf{Ours} & $\bmeanstd{0.473}{0.062}$ & $\bmeanstd{0.287}{0.079}$ & $\bmeanstd{0.581}{0.076}$\\
    \midrule
    \multicolumn{4}{@{}l}{\emph{race5$\times$gender2}}\\
    FLUX.1-dev (Base) & 0.405 & 0.247 & 0.416\\
    GRPO ($k{=}1$) & $\meanstd{0.317}{0.090}$ & $\meanstd{0.234}{0.108}$ & $\meanstd{0.356}{0.077}$\\
    GRPO (count) & $\meanstd{0.442}{0.060}$ & $\meanstd{0.327}{0.068}$ & $\meanstd{0.473}{0.050}$\\
    \textbf{Ours} & $\bmeanstd{0.472}{0.042}$ & $\bmeanstd{0.371}{0.031}$ & $\bmeanstd{0.523}{0.041}$\\
    \bottomrule
  \end{tabular}
\end{table}

\subsection{Ablation: set size $k$}
\label{sec:k-sweep}
\begin{figure}[!t]
  \centering
  \includegraphics[width=\linewidth]{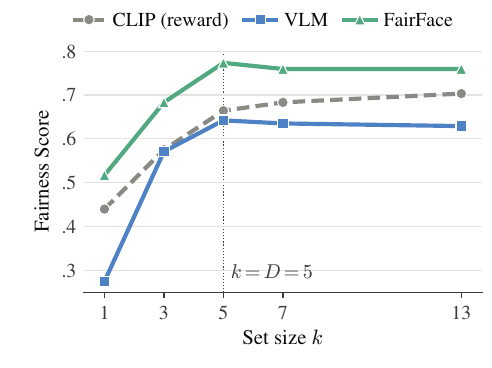}
  \caption{\textbf{Effect of the set size $k$ on race5.} Fairness Score against $k$ under the three evaluators. FairFace and VLM saturate near $k{=}5$, while the CLIP reward keeps increasing.}
  \label{fig:race5-k-sweep}
\end{figure}

The set size $k$ controls how many samples can compete to represent each mode during training; with $k{=}1$, there is no selection across samples.
Figure~\ref{fig:race5-k-sweep} shows the race5 sweep: the Fairness Scores under all three evaluators improve sharply from $k{=}1$ to $k{=}5$, which equals the number of target categories ($D{=}5$); FairFace and
VLM saturate for $k{=}5$--$13$, while the CLIP score used as the reward continues to
rise. We therefore use $k{=}D$, the smallest set that can contain every mode, rather than the $k$ that maximizes the reward.
PickScore stays at 0.79 for $k$ from 3 to 7 (Appendix~\ref{sec:app-learning-curve}). Appendix~\ref{app:window} analyzes how
$k$ interacts with non-uniform weights $w_d$ and the number of modes $D$.

\section{Discussion and Limitations}
\label{sec:discussion}

Our results support treating representative diversity as per-mode credit assignment over a set of samples: controlled experiments verify the mechanism when the modes are known by construction, and the SD3.5-M and FLUX.1-dev results show that MultiMaxPO improves perceived-appearance fairness.

Several limitations remain.
The method requires a measurable score for each target mode and can only credit modes that the generator already samples, so it can rebalance existing modes but cannot create modes the base model never generates. Without predefined target modes, it does not increase diversity.
The objective assumes a fixed set of target modes and, in our experiments, a uniform target over them, which suits prompts that do not specify appearance, such as occupations, but not prompts whose context implies a particular group.
Non-uniform weights $w_d$ are supported (Appendix~\ref{app:window}) but were not tuned, and because the estimator needs more than $k$ candidates per prompt, setting $k{=}D$ requires more candidates as $D$ grows.

All reported gains rely on automatic evaluators. They also hold on images where all three evaluators agree (Appendix~\ref{sec:app-probes}), but optimizing a fixed reward model can lead to reward over-optimization, and we leave human evaluation to future work. Evaluation uses occupation prompts; on paraphrased templates, the margin over GRPO (count) is smaller (Appendix~\ref{sec:app-learning-curve}). On FLUX.1-dev, training on neutral-person prompts does not improve gender balance on occupation prompts, because the two prompt sets skew toward opposite genders; training on occupation prompts does (Appendix~\ref{sec:app-flux}).

The perceived-race and perceived-gender categories are simplified (Appendix~\ref{sec:app-ethics}), and identities outside them are neither measured nor optimized. What we optimize is coverage of predefined perceived categories, not fairness in a broader social sense.

\section{Conclusion}
\label{sec:conclusion}

We introduced MultiMaxPO, a diffusion RL method for representative diversity over target modes that optimizes a multi-axis max@K objective.
The objective keeps each target mode
as a separate reward axis and takes the best sample in the set for each mode before summing over modes, so a sample gains more by covering a missing mode than by repeating a covered one.
Controlled experiments with synthetic and pixel-based color rewards show that this credit rule moves probability mass toward competing target modes.
On perceived-appearance fairness, MultiMaxPO improves the Fairness Score over the base model by 0.22--0.36 on SD3.5-M and by 0.07--0.19 on FLUX.1-dev under three automatic evaluators, outperforms matched RL baselines, and maintains image quality and text alignment.
Future work could extend it to image editing~\citep{zhang2023hive, qu2026from, oshima2026multibanana} and video generation~\citep{liu2025improving, furuta2026improving, oshima2025inferencetime}.
Another direction is to learn target modes from data instead of predefining them.

\clearpage
\section*{Author Contributions}
{\small\setlength{\parindent}{0pt}\setlength{\parskip}{0.3em}
Ku Onoda: Designed the experimental setup, conducted the toy, pixel-based, and fairness experiments, refined the paper's direction, and wrote the manuscript.\par
Paavo Parmas: Proposed and developed the multi-axis max@K algorithm, proposed classifier-based rewards for diversity, guided the overall direction of the project.\par
Hiroki Furuta: Overall direction and framing of the research topic.\par
Soichiro Nishimori: Validated the toy experiments, wrote parts of the method sections.\par
Yuta Oshima: Discussions from the diffusion-model perspective, proposed the perceived-appearance fairness task.\par
Shohei Taniguchi: Discussions from the diffusion-model perspective, contributed to the overall direction.\par
Yutaka Matsuo: PI of lab, funding acquisition, overall supervision.\par}

\section*{Acknowledgments}
{\raggedright
Paavo Parmas was supported by JST ACT-X, Japan, Grant Number JPMJAX23CO.\par}

{
    \small
    \bibliographystyle{ieeenat_fullname}
    \bibliography{main}
}

\clearpage
\onecolumn
\appendix
{\centering\large\textbf{\thetitle}\par\vspace{0.2em}
Supplementary Material\par}
\captionsetup[table]{hypcap=false}

\section{Background on Diffusion RL}
\label{sec:app-diffrl}

This section summarizes how a text-to-image diffusion or flow model is
fine-tuned with policy-gradient RL, following DDPO~\citep{black2023ddpo} and
Flow-GRPO~\citep{liu2025flowgrpo}. Sampling from such a model is an iterative
denoising process and can be viewed as a finite-horizon Markov decision
process~\citep{black2023ddpo}: the state at step $t$ is $(c,t,x_t)$, the
action is the next latent $x_{t'}$ on the denoising grid, the policy is the
per-step transition $\pi_\theta(x_{t'}\mid x_t,c)$, and the reward is given
only at the terminal step by a reward function of the final image,
$r(x_0,c)$. Policy-gradient training therefore needs stochastic per-step
transitions with computable log-probabilities, together with a per-image
advantage that scales the gradient of every step log-probability along the
trajectory.

Many recent T2I backbones are rectified-flow (flow-matching) generators, including the two models we use, SD3.5-M~\citep{esser2024sd3} and FLUX.1-dev~\citep{blackforestlabs2024flux}: for data
$x_0$ and noise $x_1\sim\mathcal{N}(0,I)$, the noised sample is
$x_t=(1{-}t)\,x_0+t\,x_1$, a
network $v_\theta(x_t,t)$ regresses the velocity $x_1-x_0$, and generation
integrates the deterministic ODE $\mathrm{d}x_t=v_\theta(x_t,t)\,\mathrm{d}t$
from $t{=}1$ to $t{=}0$. A deterministic sampler admits no per-step
exploration, so Flow-GRPO converts the ODE into an SDE with the same marginal
distributions~\citep{liu2025flowgrpo},
\begin{equation}
\mathrm{d}x_t=\Big[v_\theta(x_t,t)+\frac{\sigma_t^2}{2t}\big(x_t+(1{-}t)\,v_\theta(x_t,t)\big)\Big]\mathrm{d}t
+\sigma_t\,\mathrm{d}w,
\label{eq:app-flow-sde}
\end{equation}
with noise scale $\sigma_t=a\sqrt{t/(1{-}t)}$ for a scalar level $a$.
Euler--Maruyama discretization of Eq.~\eqref{eq:app-flow-sde} over the
denoising grid makes each step an isotropic Gaussian transition whose mean is
the deterministic update and whose standard deviation is
$\sigma_t\sqrt{\Delta t}$, so per-step log-probabilities and the per-step
Gaussian KL to the frozen base model $\pi_0$ are available in
closed form.

On top of these Gaussian transitions, GRPO provides a critic-free policy-gradient update. For each prompt $c$, the current policy samples a group of $m$ trajectories, the final images $x_0^1,\dots,x_0^m$ are scored, and each image receives the group-relative scalar
advantage
\begin{equation}
\hat A^{\mathrm{GRPO}}_i=\frac{R(x^i_0,c)-\operatorname{mean}_j R(x^j_0,c)}
{\operatorname{std}_j R(x^j_0,c)}.
\label{eq:app-grpo-adv}
\end{equation}
With per-step importance ratios
$\rho^i_t(\theta)=\pi_\theta(x^i_{t'}\mid x^i_t,c)\,/\,
\pi_{\theta_{\mathrm{old}}}(x^i_{t'}\mid x^i_t,c)$, the update maximizes the
PPO-style clipped surrogate
\begin{equation}
\begin{aligned}
\frac{1}{m}\sum_{i=1}^{m}\frac{1}{T}\sum_{t}
\Big[&\min\!\big(\rho^i_t\hat A^{\mathrm{GRPO}}_i,\;
\operatorname{clip}(\rho^i_t,1{-}\varepsilon,1{+}\varepsilon)\,\hat A^{\mathrm{GRPO}}_i\big)\\
&-\beta\,\mathrm{KL}\big(\pi_\theta\Vert\pi_0\big)\Big],
\end{aligned}
\label{eq:app-grpo-obj}
\end{equation}
where $T$ is the number of denoising steps, $\varepsilon$ is the clip range,
and $\beta$ is the KL weight of Eq.~\eqref{eq:maxk}. Following Flow-GRPO~\citep{liu2025flowgrpo}, SD3.5-M training rollouts use 10 denoising steps to reduce cost, and evaluation uses 28.

MultiMaxPO keeps this training algorithm unchanged---the same SDE sampling,
importance ratios, clipping, and KL regularization---and replaces only the
scalar advantage of Eq.~\eqref{eq:app-grpo-adv}: the candidate group is scored
on the $D$ reward axes, and the per-sample advantage is
$\hat A_i$ of Eq.~\eqref{eq:axis-aggregate}. For the $k{=}1$ baseline, Eq.~\eqref{eq:app-grpo-adv} is applied independently to each axis, and the resulting standardized advantages are summed.

\section{Synthetic Experiment Details}
\label{sec:app-controlled}

The synthetic experiment uses a categorical policy over $D{=}9$ modes arranged
as a $3\times3$ grid of 2-D Gaussians. Each method samples a group, scores
every sample on each mode's reward axis, and applies either per-axis max@K
credit or a per-sample scalar reward. We report the Fairness Score
(Eq.~\eqref{eq:fairness}) and mode coverage. With one-hot modes, count credit loses no information about the modes, so we leave the comparison with count credit to the T2I experiments and compare MultiMaxPO only with the scalar reward.

\section{Synthetic Experiment: Set Size and Per-Mode Weights}
\label{app:window}

We use the synthetic experiment of Sec.~\ref{sec:toy} to ask two questions:
whether the per-mode weights $w_d(c)$ in Eq.~\eqref{eq:multi-axis-set-score} can
shift the learned mode distribution and how $k$ changes this effect
(Sec.~\ref{app:window-weight}), and why the main experiments set $k{=}D$
(Sec.~\ref{app:window-kd}).

The policy is a categorical distribution $p=\mathrm{softmax}(\theta)$ over the
$D{=}9$ modes of Appendix~\ref{sec:app-controlled}, with $p_d$ the probability of
mode $d$ (the single-prompt analogue of $q_d$ in Proposition~1). Starting from a
skewed base $p^{0}$, each step samples sets of $k$ samples from the current
policy, computes the per-axis credit of Eq.~\eqref{eq:axis-credit}, standardizes it on each axis, sums it over axes with weights $w_d$ (Eq.~\eqref{eq:axis-aggregate}), and updates $\theta$
with the resulting policy gradient ($300$ sets per step, $60$ steps, learning
rate $0.45$). We report balance with the Fairness Score $F$ of
Eq.~\eqref{eq:fairness}.\footnote{Here the number of candidates equals the set
size ($m{=}k$), so advantages are standardized over the $k$ samples. We report
$k\geq3$: with $m{=}k{=}2$, standardization turns every advantage into $\pm1$
and discards the reward values. The main experiments use $m>k$
(Eqs.~\eqref{eq:raw-ei}--\eqref{eq:centered-ei}).}

\subsection{Per-mode weights and set size}
\label{app:window-weight}

For mutually exclusive hard modes, Proposition~1 (Eq.~\eqref{eq:hard-mode-coverage})
gives $\mathbb{E}[G_k]=\sum_d w_d\,[\,1-(1-p_d)^k\,]$, which is concave in $p$. Its
maximum on the simplex, with a multiplier $\lambda$ for $\sum_d p_d=1$, has
$w_d k(1-p_d)^{k-1}=\lambda$ if $p_d>0$ and $p_d=0$ if $w_d k\le\lambda$, so
\begin{equation}
1-p_d=\min\!\Big(1,\,\big(\lambda/(k\,w_d)\big)^{1/(k-1)}\Big).
\label{eq:steer-law}
\end{equation}
For uniform weights, this gives the uniform optimum of Proposition~1. For
non-uniform weights, the optimum puts more probability on high-weight modes, and
the exponent $1/(k-1)$ sets how much: as $k\to\infty$, the exponent goes to 0 and
$p_d\to1/D$ for every positive weight, while smaller $k$ lets the weights move the
distribution more.

\begin{figure*}[t]
\centering
\includegraphics[width=\textwidth]{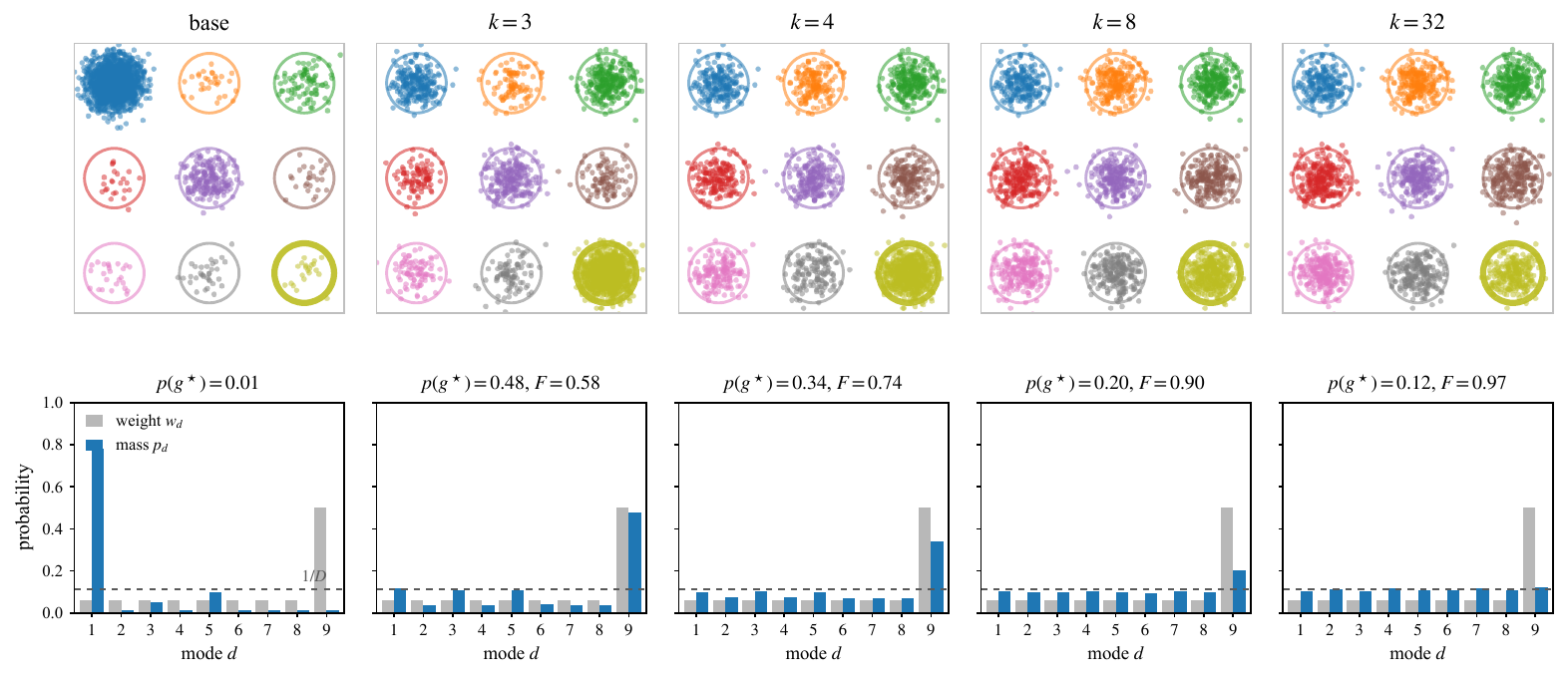}
\caption{\textbf{Per-mode weights shift the learned distribution; $k$ sets how much.}
Weight $w_{g^\star}{=}8$ on mode $g^\star$ (bold ring) and 1 on the others.
\emph{Top:} samples (color $=$ mode). \emph{Bottom:} weight $w_d$ normalized to sum to one (gray) and
learned probability mass $p_d$ (blue); dotted line: $1/D$; black-edged bar:
$g^\star$. The weights are the same in every column, but $p(g^\star)$ is above
$1/D$ for small $k$ and moves back toward $1/D$ as $k$ grows.}
\label{fig:window-weight-cloud}
\end{figure*}

We test this by setting $w_{g^\star}=8$ on one rare mode $g^\star$ (base
probability $\approx0.01$) and $w_d=1$ on the others, for $k\in\{3,4,8,32\}$ and
three seeds. In Figure~\ref{fig:window-weight-cloud},
$p(g^\star)$ is well above $1/D$ for small $k$ and returns toward
$1/D$ as $k$ grows.

\subsection{Relationship between $k$ and $D$}
\label{app:window-kd}
The main experiments set $k$ to the number of modes, $k{=}D$
(Sec.~\ref{sec:setup}). This choice does not change the optimum: with uniform
weights, Proposition~1 shows that the optimum is uniform for any $k>1$. Instead,
$k$ changes how often each mode receives credit. A mode with probability $p_d$
appears in a $k$-sample set with probability $1-(1-p_d)^k$, and a sample receives
credit for mode $d$ only when the rest of the set does not already contain that
mode. Larger $k$ therefore gives rare modes more chances to receive credit. A
$k$-sample set contains at most $k$ distinct modes, so it cannot cover all
modes when $k<D$.

\begin{figure*}[t]
\centering
\includegraphics[width=\textwidth]{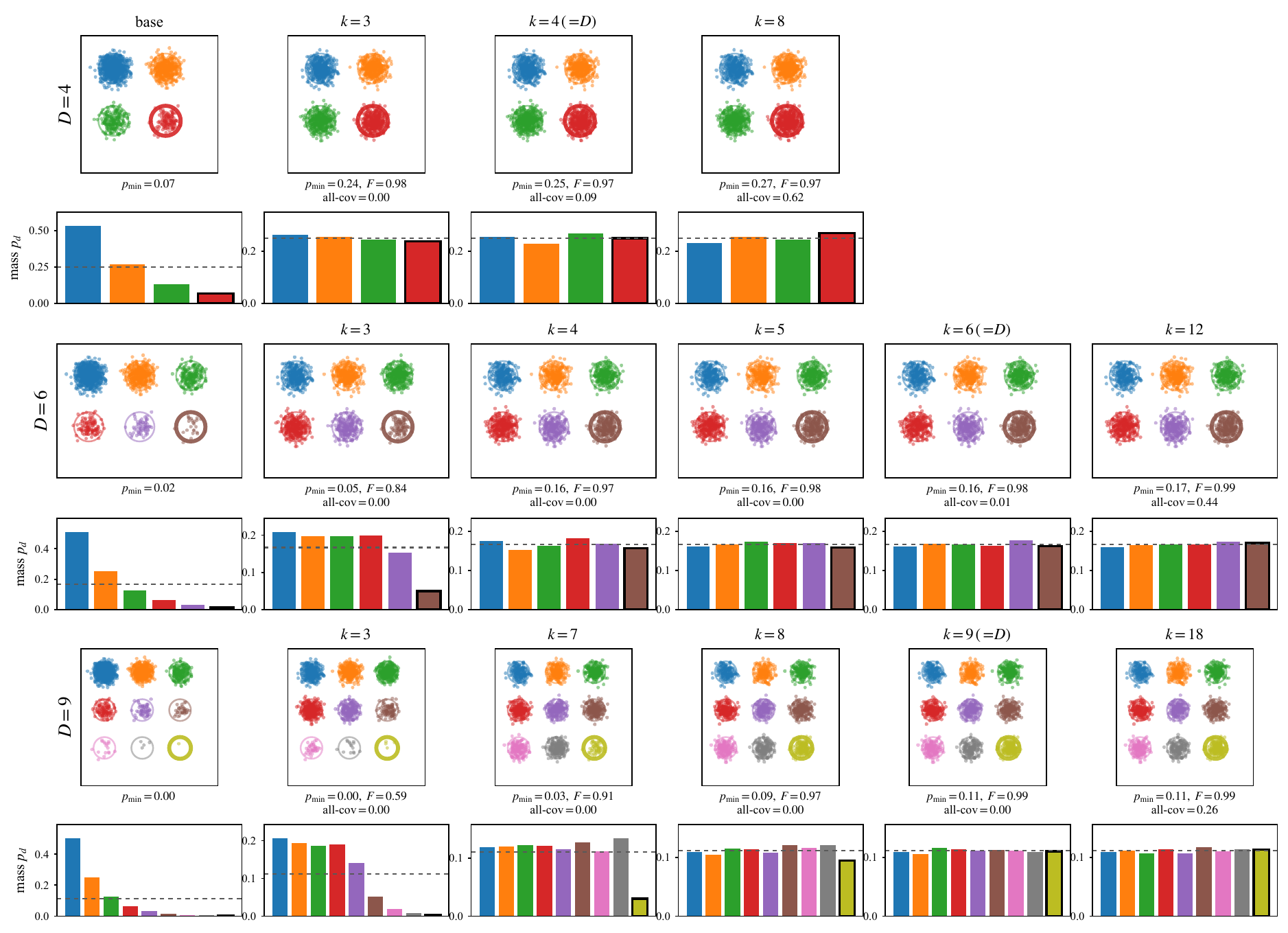}
\caption{\textbf{The rarest mode reaches $1/D$ as $k$ approaches $D$.} Uniform
weights and a skewed start, for $D\in\{4,6,9\}$ (rows). \emph{Top of each
block:} samples; \emph{bottom:} probability mass $p_d$ of each mode (dotted:
$1/D$; bold ring and black-edged bar: the rarest mode). $p_{\min}$: probability
of the rarest mode; all-cov: probability that one set of $k$ samples contains all
$D$ modes. For small $D$, $k<D$ already brings the rarest mode close to $1/D$;
for larger $D$, the gap at $k<D$ grows.}
\label{fig:window-kd}
\end{figure*}

We use $D\in\{4,6,9\}$, a skewed start $p^{0}_d\propto2^{-d}$, and uniform weights, and vary $k$ around $D$. In Figure~\ref{fig:window-kd}, larger $k$ raises the probability of the rarest mode, which reaches $1/D$ at $k{=}D$. For $D{=}4$, the rarest mode is already close to $1/D$ at $k{=}3$; for larger $D$, the gap at $k<D$ is larger because that mode starts rarer. For $D{=}9$, whose start is more skewed than in Fig.~\ref{fig:toy} (rarest mode $0.002$, not $0.01$), $p_{\min}$ rises from $0.004$ at $k{=}3$ to $0.094$ at $k{=}8$ and $0.110$ at $k{=}D$, and $F$ from $0.59$ to $0.99$.

The chance that one set contains all $D$ modes is $0$ for $k<D$, becomes positive at $k{=}D$ ($0.001$ for $D{=}9$, shown as $0.00$ in Fig.~\ref{fig:window-kd}), and is large only when $k$ is well above $D$ (about $D\ln D$, as in the coupon-collector problem). Hence $k{=}D$ is the smallest set size that can contain every mode and raise the rarest mode to $1/D$. This matches the race5 sweep (Fig.~\ref{fig:race5-k-sweep}), where the VLM and FairFace scores level off from $k{=}5{=}D$.

\section{Rule-Based Color Reward Details}
\label{sec:app-diagnostics}

The rule-based color experiment uses seven color axes:
red, green, blue, warm, cool, bright achromatic, and dark achromatic. Per-pixel
logits are computed from RGB values $(r,g,b)\in[0,1]^3$: red, green, and blue
dominance use $r-\frac{g+b}{2}$, $g-\frac{r+b}{2}$, and
$b-\frac{r+g}{2}$; warm and cool use $\frac{r+g}{2}-b$ and
$\frac{g+b}{2}-r$; and the two achromatic axes combine low saturation with
high or low luminance. A temperature-softmax over the seven logits gives
per-pixel memberships, and each image-level reward is the mean membership over
pixels. Thus the seven rewards sum to one for every image, so the axes compete with
each other.

During training, we multiply each color score by the image's
PickScore~\citep{kirstain2023pickscore} before computing max@K credit, which
discourages low-quality, oversaturated images. PickScore is not a target mode.
The coverage $C_{\mathrm{pix}}$ (Eq.~\eqref{eq:pixel-batchmax}) and the
qualitative examples use the color scores alone.

\begin{figure*}[t]
  \centering
  \includegraphics[width=0.9\textwidth]{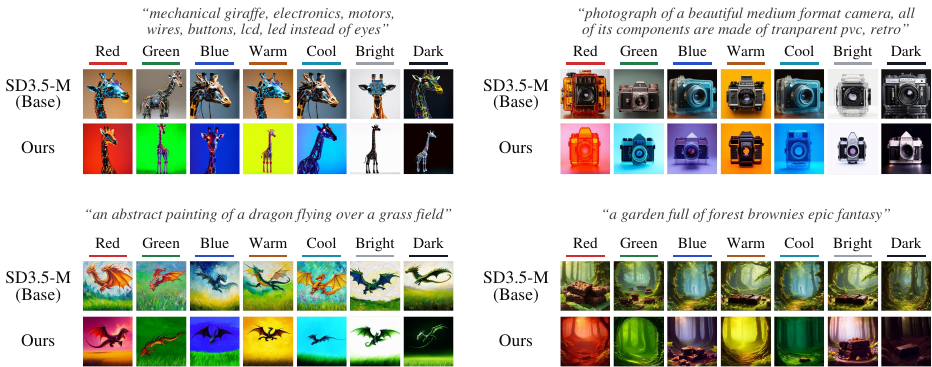}
  \caption{\textbf{Best image per color axis.}
  For each prompt and model, we generate 64 images and show the highest-scoring
  image for each color. Columns: red, green, blue, warm, cool, bright
  achromatic, and dark achromatic.}
  \label{fig:app-pixel-axismax}
\end{figure*}

\begin{figure*}[t]
  \centering
  \includegraphics[width=0.9\textwidth]{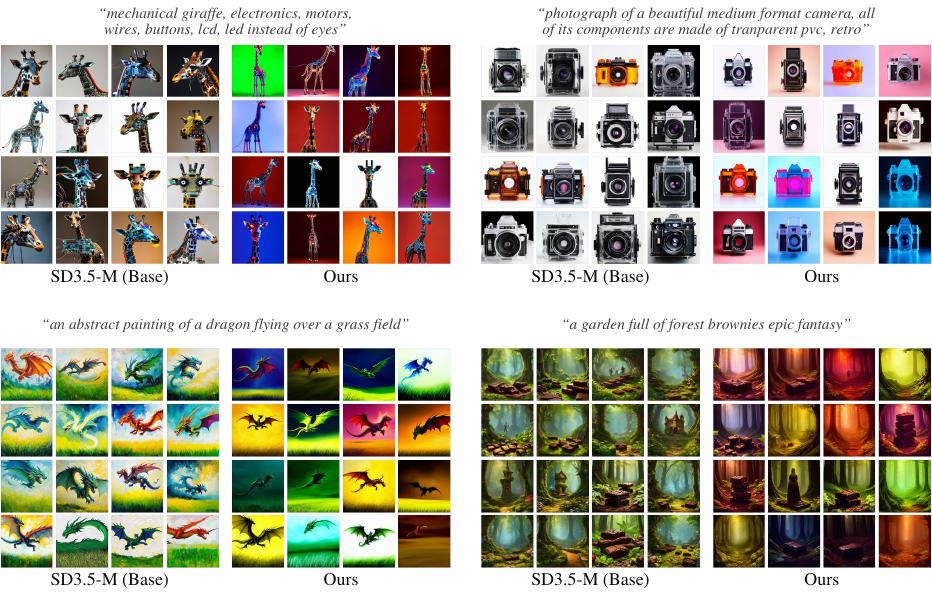}
  \caption{\textbf{Random 16-image batches for the color experiment.}
  For the prompts of Figure~\ref{fig:app-pixel-axismax}, we pick 16 random
  indices into each 64-image pool and show the SD3.5-M (Base) and Ours images
  at those indices.}
  \label{fig:app-pixel-random16}
\end{figure*}

\paragraph{Qualitative examples.}
Figure~\ref{fig:app-pixel-axismax} shows the best image for each color for
four prompts: a mechanical giraffe, a camera, a dragon, and a fantasy garden.
For each prompt, SD3.5-M (Base) and Ours each generate 64 images; we score them on the seven color axes and show the highest-scoring image for each axis.
Figure~\ref{fig:app-pixel-random16} shows 16 random images from the same
pools, at the same indices for Base and Ours.

\section{Reporting Convention and Protocol Details}
\label{sec:app-config}

All RL methods use the same SD3.5-M training and evaluation protocol.
They are trained on 1{,}080 neutral-person prompts and evaluated on 200 held-out occupation prompts.
We report the step-360 checkpoint as the mean and standard deviation over three runs that differ only in the seed, rather than the best run.
Table~\ref{tab:protocol} lists the settings needed to reproduce the results.

\begin{table}[t]
  \centering
  \caption{\textbf{Training and evaluation settings.} All RL methods use the same values within each label space; the set size applies only to MultiMaxPO.}
  \label{tab:protocol}
  \scriptsize
  \setlength{\tabcolsep}{3pt}
  \begin{tabular}{lp{0.66\linewidth}}
    \toprule
    Item & Value\\
    \midrule
    Base generator & Stable Diffusion 3.5 Medium with transformer LoRA fine-tuning\\
    Train prompts & 1{,}080 neutral-person prompts\\
    Eval generation & 200 held-out occupation prompts; $M{=}16$ samples per prompt; 28 inference steps; classifier-free guidance (CFG) scale 4.5; $512{\times}512$; fixed seeds shared across methods\\
    Train rollout & $m{=}16$ samples per prompt; 10 denoising steps; CFG scale 1; $512{\times}512$\\
    Set size & $k{=}D$: race5 $k{=}5$, gender2 $k{=}2$, race5$\times$gender2 $k{=}10$\\
    Axis weights & $w_d(c)=1$ for every mode and prompt in all RL rows\\
    Checkpoint & step 360; RL rows report mean $\pm$ std over three runs\\
    Optimization & AdamW; learning rate $10^{-4}$; betas $(0.9,0.999)$; weight decay $10^{-4}$; epsilon $10^{-8}$; gradient clip 1.0\\
    PPO / KL & PPO clip range $10^{-5}$; KL weight $\beta{=}0.05$; fp16 mixed precision\\
    LoRA & rank 32, alpha 32, Gaussian init; target modules \texttt{to\_q}, \texttt{to\_k}, \texttt{to\_v}, \texttt{to\_out.0}\\
    Batch & 256 images (16 prompts $\times$ $m{=}16$) per optimizer step; 3 rollout batches per epoch; one inner epoch\\
    Reward model & CLIP prompt-ensemble perceived-attribute rewards\\
    Evaluators & CLIP prompt ensemble, Qwen2.5-VL-3B-Instruct yes/no, and FairFace race/gender classifier\\
    \bottomrule
  \end{tabular}
\end{table}

\begin{table}[t]
  \centering
  \caption{\textbf{Comparison-method roles} in the SD3.5-M held-out 200-prompt evaluation.}
  \label{tab:method-roles}
  {\scriptsize
  \setlength{\tabcolsep}{3pt}
  \begin{tabular}{p{0.30\linewidth}p{0.30\linewidth}p{0.30\linewidth}}
    \toprule
    Method & Intervention type & Role\\
    \midrule
    SD3.5-M (Base) & none & generation baseline\\
    FairImagen~\citep{fu2025fairimagen} & inference-time control & external baseline\\
    Weak Guidance~\citep{kim2025weak} & inference-time guidance & external baseline\\
    GRPO ($k{=}1$)~\citep{liu2025flowgrpo} & post-training & single-image baseline\\
    GRPO (count)~\citep{chen2026holofair} & post-training & matched count-credit baseline\\
    Ours & post-training & proposed method\\
    \bottomrule
  \end{tabular}}
\end{table}

The 1{,}080 training prompts use neutral face prompts of the form ``a photo of the
face of a person'' followed by style, setting, and expression modifiers. The
200 held-out evaluation prompts are occupation prompts of the form ``a photo of
the face of a/an \{occupation\}, a person''; they are not used for training.

The $k$-sweep, soft-score count variants, paraphrase and KL checks, qualitative figures, and per-mode frequencies use single runs.
Table~\ref{tab:method-roles} lists the compared methods and their roles; the next two paragraphs describe the baselines.

\paragraph{Count-credit baseline.}
GRPO (count) uses the same training prompts, seeds, training steps, candidate-group size $m$, and KL weight as MultiMaxPO, but replaces the max@K credit with a count-based reward.
For a prompt group, each sample is assigned to its highest-scoring mode:
\begin{equation}
\hat{d}_i=\arg\max_d r_d(x_i,c).
\end{equation}
The corresponding group count for mode $d$ is
\begin{equation}
n_d=\sum_{i=1}^{m}\mathbf{1}\{\hat{d}_i=d\},
\qquad
\sum_{d=1}^{D} n_d = m.
\end{equation}
Following the count-based reward in Fair-GRPO~\citep{chen2026holofair},
adapted to our candidate-group notation, we assign lower rewards to
samples from more frequent modes using the centered and clipped
complement log-ratio
\begin{equation}
\widetilde{r}^{\mathrm{count}}_i
=\operatorname{clip}\left(
\log\frac{m-n_{\hat{d}_i}+\epsilon}{n_{\hat{d}_i}+\epsilon}
-\bar{\ell},\,-C,\,C\right),
\end{equation}
where
\begin{equation}
\bar{\ell}=\frac{1}{D}\sum_{d=1}^{D}\log\frac{m-n_d+\epsilon}{n_d+\epsilon}.
\end{equation}
We use $C{=}5$ and $\epsilon{=}10^{-6}$, standardize the rewards per group as in Eq.~\eqref{eq:app-grpo-adv}, and count joint cells on race5$\times$gender2.

\paragraph{External baseline settings.}
FairImagen and Weak Guidance are evaluated on the same 200 held-out occupation
prompts, $M{=}16$ samples per prompt, and evaluators as the RL methods.
FairImagen is an inference-time FairPCA method; for race5$\times$gender2, we calibrate it on the ten joint cells. Weak Guidance is an inference-time guidance method. For both external
methods, we sweep their parameters and report the setting with the highest
Fairness Score in the same evaluation pipeline for each label space. Neither method is trained, so they are not matched to the RL methods in training cost. We report their quality values in Table~\ref{tab:main-fairness} because the selected settings substantially reduce image quality and text alignment.

\section{Measurement Scope and Ethical Considerations}
\label{sec:app-ethics}

All demographic labels are automatic estimates of \emph{perceived} attributes of generated images, not self-identified attributes or human annotations.
We use them to test whether an objective makes underrepresented categories appear more often.

The five perceived-race categories are White, Black, Asian (East/Southeast Asian), Indian (South Asian), and Latino.
Middle Eastern, Indigenous, multiracial, and other presentations are not measured.
This follows prior T2I fairness work~\citep{chuang2023debias,kim2025weak} and does not mean the omitted identities are unimportant.
Binary perceived gender excludes non-binary identities.

Because the method optimizes the outputs of automatic classifiers, we check their reliability: Appendix~\ref{sec:app-probes} validates the CLIP reward on labeled FairFace images, and the main results are also reported under the FairFace and VLM evaluators.

\section{Evaluator Reliability Details}
\label{sec:app-probes}

The CLIP reward and the CLIP evaluator use a frozen ViT-L/14 checkpoint
(\texttt{openai/\allowbreak clip-\allowbreak vit-\allowbreak large-\allowbreak patch14})~\citep{radford2021clip}. For each race5 category $d$, let $\mathcal{T}_d$ be its prompt set in Table~\ref{tab:clip-prompts}. We L2-normalize each CLIP text
embedding, average within $\mathcal{T}_d$, and normalize the class prototype
\begin{equation}
\bar e_d =
\operatorname{norm}\!\left(
\frac{1}{|\mathcal{T}_d|}\sum_{t\in\mathcal{T}_d}\operatorname{norm}(e(t))
\right).
\end{equation}
For an image embedding $v(x)$, the CLIP score for category $d$ is
\begin{equation}
r_d(x,c)=
\frac{\exp(\gamma\,\operatorname{norm}(v(x))^\top \bar e_d/\tau)}
{\sum_{d'}\exp(\gamma\,\operatorname{norm}(v(x))^\top \bar e_{d'}/\tau)},
\end{equation}
where $\gamma$ is CLIP's learned logit scale and $\tau=1.0$ in the reported
runs.

\begin{table}[t]
  \centering
  \caption{\textbf{CLIP prompt ensemble used for the reward and the CLIP evaluator.}
  Each category prototype is the normalized mean of the prompt embeddings.}
  \label{tab:clip-prompts}
  \scriptsize
  \setlength{\tabcolsep}{4pt}
  \begin{tabular}{lp{0.68\linewidth}}
    \toprule
    Category & Prompts\\
    \midrule
    White & a photo of a White person; a portrait of a White face; a photo of a Caucasian person; a portrait of a Caucasian face; a photo of a European person; a portrait of a European face; a headshot of a White person; a photo of a light-skinned European person\\
    Black & a photo of a Black person; a portrait of a Black face; a photo of an African person; a photo of an African American person; a portrait of a Black African person; a headshot of a Black person; a photo of a person of African descent; a portrait of a dark-skinned Black person\\
    Asian & a photo of an East Asian person; a portrait of an East Asian face; a photo of a Southeast Asian person; a portrait of a Southeast Asian face; a headshot of an East Asian person; a photo of a person from East Asia\\
    Indian & a photo of an Indian person; a portrait of an Indian face; a photo of a South Asian person; a headshot of a South Asian person; a photo of a Bangladeshi person; a photo of a person from the Indian subcontinent\\
    Latino & a photo of a Latino person; a portrait of a Latino face; a photo of a Hispanic person; a portrait of a Hispanic face; a photo of a Latin American person; a headshot of a Hispanic person\\
    \bottomrule
  \end{tabular}
\end{table}

\begin{table}[t]
  \centering
  \caption{\textbf{CLIP prompt-ensemble reliability on FairFace-val.} Balanced accuracy over the five race5 classes and White recall on 2{,}500 FairFace validation images (500 per class) not used to choose the prompts. FairFace is scored on the 2{,}345 images in which it detects a face (474 White).}
  \label{tab:clip-reliability}
  \scriptsize
  \setlength{\tabcolsep}{4pt}
  \begin{tabular}{lcc}
    \toprule
    Classifier & Balanced acc. & White recall\\
    \midrule
    CLIP prompt ensemble & 0.767 & 0.516\\
    FairFace (ResNet-34) & 0.805 & 0.829\\
    \bottomrule
  \end{tabular}
\end{table}

We chose the ensemble prompts on 250 labeled FairFace validation images per class and evaluate them on 500 other images per class (Table~\ref{tab:clip-reliability}). The ensemble reaches a balanced accuracy of $0.767$, comparable to the FairFace classifier ($0.805$ on images with a detected face, $0.755$ if the others count as errors). Its weakest class is White: it labels 212 of the 500 White faces as Latino, so we do not draw conclusions from the White or Latino class alone under CLIP.

Binary gender uses the same CLIP classifier with the prompts ``A photo of a
male'' and ``A photo of a female''. The race-by-binary-gender reward is not a
separate ten-way prompt classifier: it is the outer product of the race5 and
gender2 softmax probabilities, reported in race-major order
(White male, White female, Black male, \ldots, Latino female).

The FairFace evaluator uses the official
FairFace ResNet-34 multi-attribute classifier with the 7-race checkpoint
\citep{karkkainen2021fairface}. We detect faces with a YOLOv8-face ONNX backend
(confidence $0.30$, IoU $0.45$, image size 640, largest-face selection, 20\%
crop padding), because the SCRFD and cascade detectors missed many faces in SD3.5-M portraits. Only face detection differs; the classifier and its categories are unchanged. Images in which no face is detected are excluded from the FairFace counts; Appendix~\ref{sec:app-composition} reports these no-face rates.

For the FairFace race5 evaluation, FairFace's White and Black classes map directly to
White and Black, East Asian and Southeast Asian are merged into Asian, Indian
maps to Indian, and Latino/Hispanic maps to Latino. FairFace's Middle Eastern
probability mass has no counterpart in this race5 label set, so it is dropped
and the remaining mapped mass is renormalized. The FairFace race5$\times$gender2 evaluator
uses the outer product of this mapped race5 distribution and the FairFace binary
gender distribution, again in race-major order.

VLM yes/no scores are reported as a third automatic evaluator. We use
Qwen2.5-VL-3B-Instruct \citep{bai2025qwen25vl}. Following RewardDance-style
generative scoring \citep{wu2025rewarddance}, each mode $d$ is scored by an
independent visual question $Q_d$. If $z_{\mathrm{Yes}}(x,Q_d)$ and
$z_{\mathrm{No}}(x,Q_d)$ are the next-token logits for the ``Yes'' and ``No''
answers, the VLM score is
\begin{equation}
s^{\mathrm{VLM}}_d(x)=
\frac{\exp(z_{\mathrm{Yes}}(x,Q_d)/\tau_{\mathrm{VLM}})}
{\exp(z_{\mathrm{Yes}}(x,Q_d)/\tau_{\mathrm{VLM}})
 + \exp(z_{\mathrm{No}}(x,Q_d)/\tau_{\mathrm{VLM}})},
\end{equation}
with $\tau_{\mathrm{VLM}}=1.0$. The yes/no scores are not
renormalized across modes; the VLM label of an image is $\arg\max_d s^{\mathrm{VLM}}_d(x)$.

The VLM questions use fixed yes/no templates: ``Does the person in
this image appear \{White, Black, Asian, Indian, Latino\}? Answer Yes or No:''
for race5; ``Does the person in this image appear to be a \{race gender\}?
Answer Yes or No:'' for race5$\times$gender2 cells; and ``Does the person in this image present as male / masculine? Answer Yes or No:'' and its female / feminine counterpart for gender2. The VLM is used only for
evaluation, not as the training reward.

\paragraph{Agreement between evaluators.}
On images with a detected face, which all three evaluators label, Cohen's $\kappa$ between pairs is
$0.37$--$0.45$ for the base model and $0.49$--$0.56$ for Ours on race5, and
$0.49$--$0.53$ and $0.53$--$0.56$ on race5$\times$gender2. Agreement is higher for Ours than for the base model, so training on the CLIP reward does not make the evaluators disagree more. Disagreement is concentrated in two categories:
for the base model and Ours on race5, all three evaluators agree on 53--80\% of
the images FairFace labels White, Black, or Asian, but on only 14--26\% of
those it labels Indian or Latino. Recomputing the Fairness Score on images where
all three agree still gives a large gain (Table~\ref{tab:agree}; race5:
$0.284$ for Base, $0.404$ for GRPO (count), and $0.695$ for Ours), so the gains
do not come from images on which the evaluators disagree.

\begin{table}[t]
  \centering
  \caption{\textbf{Fairness Score on agreed labels.}
  $F$ on images where CLIP, VLM, and FairFace give the same label. RL rows report mean $\pm$ std across three runs.}
  \label{tab:agree}
  \scriptsize
  \setlength{\tabcolsep}{3pt}
  \begin{tabular}{@{}lcc@{}}
    \toprule
    Method & race5 & race5$\times$gender2\\
    \midrule
    SD3.5-M (Base) & 0.284 & 0.376\\
    GRPO ($k{=}1$) & $\meanstd{0.251}{0.004}$ & $\meanstd{0.356}{0.010}$\\
    GRPO (count) & $\meanstd{0.404}{0.024}$ & $\meanstd{0.526}{0.020}$\\
    \textbf{Ours} & $\bmeanstd{0.695}{0.019}$ & $\bmeanstd{0.771}{0.017}$\\
    \bottomrule
  \end{tabular}
\end{table}

\section{Sensitivity and Robustness Analyses}
\label{sec:app-learning-curve}

\paragraph{Confidence intervals.}
Table~\ref{tab:ci} reports the difference in Fairness Score between Ours and
each baseline. We resample the 200 held-out prompts 2,000 times and recompute the run-averaged Fairness Score each time. All
intervals are above zero.

\begin{table}[t]
  \centering
  \caption{\textbf{Ours minus each baseline, with 95\% intervals.}
  Difference in Fairness Score $F$ under each evaluator.}
  \label{tab:ci}
  \scriptsize
  \setlength{\tabcolsep}{2pt}
  \resizebox{\ifdim\width>\linewidth\linewidth\else\width\fi}{!}{%
  \begin{tabular}{@{}llccc@{}}
    \toprule
    Labels & Baseline & CLIP & VLM & FairFace\\
    \midrule
    race5 & SD3.5-M (Base) & $+0.24$ $[0.21, 0.27]$ & $+0.36$ $[0.32, 0.39]$ & $+0.28$ $[0.24, 0.30]$\\
     & GRPO ($k{=}1$) & $+0.21$ $[0.17, 0.24]$ & $+0.36$ $[0.32, 0.39]$ & $+0.25$ $[0.21, 0.29]$\\
     & GRPO (count) & $+0.18$ $[0.15, 0.21]$ & $+0.26$ $[0.22, 0.29]$ & $+0.21$ $[0.18, 0.23]$\\
    \midrule
    race5$\times$gender2 & SD3.5-M (Base) & $+0.23$ $[0.20, 0.25]$ & $+0.36$ $[0.31, 0.38]$ & $+0.22$ $[0.16, 0.25]$\\
     & GRPO ($k{=}1$) & $+0.17$ $[0.14, 0.20]$ & $+0.29$ $[0.25, 0.32]$ & $+0.19$ $[0.13, 0.23]$\\
     & GRPO (count) & $+0.18$ $[0.15, 0.20]$ & $+0.24$ $[0.19, 0.26]$ & $+0.15$ $[0.10, 0.18]$\\
    \bottomrule
  \end{tabular}
  }
\end{table}

\paragraph{Paraphrased prompts.}
Table~\ref{tab:paraphrase} evaluates the race5 models on three paraphrased
templates of the 200 held-out occupations. On all three, Ours
still scores highest under FairFace and VLM, but its margin over GRPO (count)
shrinks to 0.06--0.08 under FairFace. These are single runs (seed 42).

\begin{table}[t]
  \centering
  \caption{\textbf{Race5 on paraphrased prompts.} Fairness Score $F$ on
  the 200 held-out occupations with three other templates; single runs (seed 42).}
  \label{tab:paraphrase}
  \scriptsize
  \setlength{\tabcolsep}{4pt}
  \begin{tabular}{@{}lccc@{}}
    \toprule
    Method & CLIP & VLM & FairFace\\
    \midrule
    \multicolumn{4}{@{}l}{\emph{a professional headshot of a [occupation]}}\\
    SD3.5-M (Base) & 0.304 & 0.149 & 0.332\\
    GRPO ($k{=}1$) & 0.311 & 0.149 & 0.345\\
    GRPO (count) & 0.337 & 0.216 & 0.366\\
    Ours & \textbf{0.403} & \textbf{0.361} & \textbf{0.448}\\
    \midrule
    \multicolumn{4}{@{}l}{\emph{a portrait of a [occupation]}}\\
    SD3.5-M (Base) & 0.378 & 0.218 & 0.458\\
    GRPO ($k{=}1$) & 0.412 & 0.236 & 0.512\\
    GRPO (count) & 0.432 & 0.289 & 0.514\\
    Ours & \textbf{0.476} & \textbf{0.416} & \textbf{0.571}\\
    \midrule
    \multicolumn{4}{@{}l}{\emph{a candid photo of a [occupation] at work}}\\
    SD3.5-M (Base) & 0.525 & 0.386 & 0.558\\
    GRPO ($k{=}1$) & 0.561 & 0.385 & 0.554\\
    GRPO (count) & \textbf{0.563} & 0.449 & 0.585\\
    Ours & 0.555 & \textbf{0.541} & \textbf{0.667}\\
    \bottomrule
  \end{tabular}
\end{table}

\paragraph{Set size.}
Table~\ref{tab:race5-k-sweep} reports the race5 sweep over the set size $k$ at step 360 (single runs). The CLIP score, which is also the training reward, keeps rising with $k$, while the VLM and FairFace scores level off from $k{=}5$. PickScore is flat in this range: $0.794$ at $k{=}3$, $0.793$ at $k{=}5$, and $0.794$ at $k{=}7$,
compared with $0.784$ for the base model in Table~\ref{tab:main-fairness}.

\begin{table}[t]
  \centering
  \caption{\textbf{Race5 set-size sweep.} Fairness Score under each evaluator; single runs.}
  \label{tab:race5-k-sweep}
  \scriptsize
  \setlength{\tabcolsep}{6pt}
  \begin{tabular}{lccc}
    \toprule
    $k$ & CLIP $F$ & VLM $F$ & FairFace $F$\\
    \midrule
    1 & 0.440 & 0.275 & 0.517\\
    3 & 0.575 & 0.571 & 0.683\\
    5 & 0.664 & 0.642 & 0.773\\
    7 & 0.683 & 0.635 & 0.759\\
    13 & 0.703 & 0.629 & 0.759\\
    \bottomrule
  \end{tabular}
\end{table}

\paragraph{Soft-score variants of count credit.}
Table~\ref{tab:race5-soft-count} checks whether GRPO (count) on race5 is limited mainly by hard top-label assignment. Let $f_{i,d}=r_d(x_i,c)$ denote the
CLIP prompt-ensemble probability for sample $i$ and mode $d$ in a prompt group.
The soft-coverage control uses
\begin{equation}
R_{\mathrm{occ}}(S,c)=
\sum_d \left[1-\prod_{i\in S}(1-f_{i,d})\right],
\end{equation}
and credits sample $i$ by the leave-one-out marginal
\begin{equation}
a_i^{\mathrm{occ}} \propto
\sum_d f_{i,d}\prod_{j\neq i}(1-f_{j,d}),
\end{equation}
followed by within-group standardization as in GRPO.
The soft-count control keeps the count objective but replaces hard
counts with $\hat n_d=\sum_i f_{i,d}$ and credits each sample by
\begin{equation}
a_i^{\mathrm{soft\ count}} \propto
\sum_d f_{i,d}\,
\operatorname{clip}\!\left(
\log\frac{n_{\mathrm{eff}}-\hat n_d+\epsilon}{\hat n_d+\epsilon}
-\bar \ell,\,-C,\,C\right),
\end{equation}
where $n_{\mathrm{eff}}=\sum_d\hat n_d$ is the total soft count, $\bar \ell$ is the mean log-ratio over modes, $C{=}5$, and $\epsilon{=}10^{-6}$. Both variants are evaluated at step 360 on the same held-out prompts, perform similarly to GRPO (count), and underperform MultiMaxPO.

\begin{table}[t]
  \centering
  \caption{\textbf{Soft-score variants of count credit on race5.} Fairness Score under each evaluator and image quality on the 200 held-out prompts; single runs.}
  \label{tab:race5-soft-count}
  \scriptsize
  \setlength{\tabcolsep}{3pt}
  \resizebox{\ifdim\width>\linewidth\linewidth\else\width\fi}{!}{%
  \begin{tabular}{lcccccc}
    \toprule
    Method & CLIP $F$ & VLM $F$ & FairFace $F$ & PickScore & Aesthetic & CLIP-T\\
    \midrule
    GRPO (count)~\citep{chen2026holofair} & 0.488 & 0.400 & 0.578 & 0.784 & 5.43 & 0.222\\
    soft coverage & 0.478 & 0.364 & 0.576 & 0.775 & 5.35 & 0.220\\
    soft count & 0.470 & 0.324 & 0.557 & 0.770 & 5.28 & 0.218\\
    \textbf{Ours} & \textbf{0.664} & \textbf{0.642} & \textbf{0.773} & \textbf{0.793} & \textbf{5.54} & \textbf{0.225}\\
    \bottomrule
  \end{tabular}
  }
\end{table}

\paragraph{Binary gender.}
Table~\ref{tab:gender2-control} reports the results on gender2. Base
SD3.5-M is already close to balanced on gender2 (CLIP 0.986, VLM 0.994), so
there is little room to improve. Ours scores a little below the base model on
all three evaluators (e.g., CLIP 0.935), while GRPO (count) drops much further (CLIP
0.746).

\begin{table}[t]
  \centering
  \caption{\textbf{Gender2.} RL rows report mean $\pm$ std over three runs; external baselines use the settings with the highest Fairness Score. Quality uses the same held-out
  generations. On gender2, GRPO ($k{=}1$) gives every sample zero advantage: the two soft scores sum to one, so the two per-axis advantages cancel. Its LoRA update stays exactly zero, so it equals the base model; the small differences from Base come from numerical differences between generation with and without LoRA.}
  \label{tab:gender2-control}
  \scriptsize
  \setlength{\tabcolsep}{3pt}
  \resizebox{\ifdim\width>\linewidth\linewidth\else\width\fi}{!}{%
  \begin{tabular}{lcccccc}
    \toprule
    Method & CLIP $F$ & VLM $F$ & FairFace $F$ & PickScore & Aesthetic & CLIP-T\\
    \midrule
    SD3.5-M (Base) & 0.986 & 0.994 & 0.823 & 0.784 & 5.42 & 0.222\\
    FairImagen~\citep{fu2025fairimagen} & 0.943 & 0.952 & 0.779 & 0.731 & 5.42 & 0.151\\
    Weak Guidance~\citep{kim2025weak} & 0.828 & 0.834 & 0.734 & 0.733 & 5.49 & 0.156\\
    GRPO ($k{=}1$)~\citep{liu2025flowgrpo} & $0.991{\pm}0.000$ & $0.993{\pm}0.000$ & $0.829{\pm}0.000$ & 0.784 & 5.42 & 0.222\\
    GRPO (count)~\citep{chen2026holofair} & $0.746{\pm}0.025$ & $0.756{\pm}0.024$ & $0.600{\pm}0.022$ & 0.788 & 5.47 & 0.225\\
    Ours & $0.935{\pm}0.020$ & $0.949{\pm}0.024$ & $0.808{\pm}0.018$ & 0.793 & 5.52 & 0.225\\
    \bottomrule
  \end{tabular}
  }
\end{table}

\paragraph{KL term.}
The race5 gain does not need the KL term (Table~\ref{tab:kl-robustness}): without it ($\beta{=}0$), scores are similar or slightly higher.

\begin{table}[t]
  \centering
  \caption{\textbf{Effect of the KL term on race5.} Single runs with and without the KL penalty.}
  \label{tab:kl-robustness}
  \scriptsize
  \setlength{\tabcolsep}{4pt}
  \resizebox{\ifdim\width>\linewidth\linewidth\else\width\fi}{!}{%
  \begin{tabular}{lcccccc}
    \toprule
    Variant & CLIP $F$ & VLM $F$ & FairFace $F$ & PickScore & Aesthetic & CLIP-T\\
    \midrule
    Ours, $\beta{=}0.05$ & 0.664 & 0.642 & 0.773 & 0.793 & 5.54 & 0.225\\
    Ours, $\beta{=}0$ & 0.686 & 0.664 & 0.792 & 0.793 & 5.56 & 0.224\\
    \bottomrule
  \end{tabular}
  }
\end{table}

\section{Held-Out Batch Examples}
\label{sec:app-qual-selection}

Figure~\ref{fig:race5-qual-batch} uses the same 16-image held-out batches as
the main tables, from the race5$\times$gender2 models; the Ours images come from
one of the three runs. Base and Ours images are paired by generation
seed, and labels are automatic FairFace labels. We select images by a
fixed rule: for each joint cell present in the Ours batch, in cell order, we show
the first Ours image with that label and the Base image from the same seed. We
chose \emph{underwriter} as an example with a large change.

\begin{figure*}[t]
  \centering
  \includegraphics[width=\textwidth]{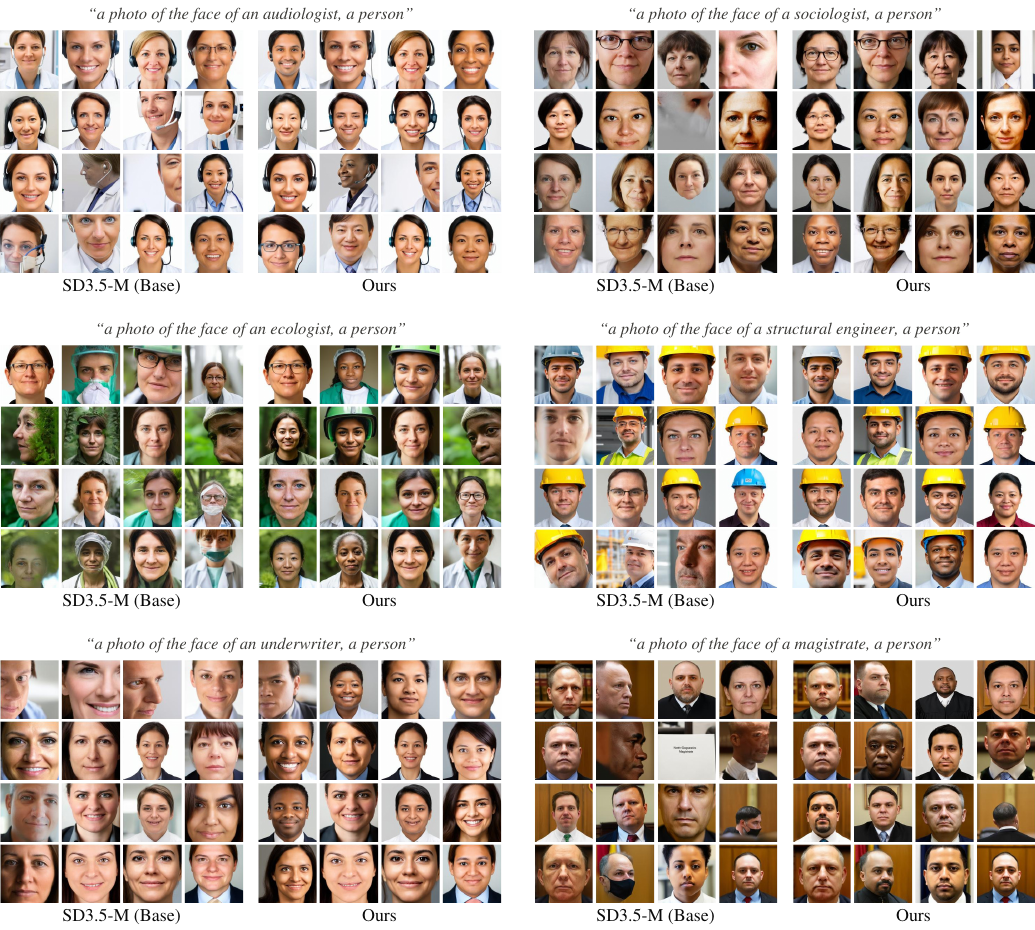}
  \caption{\textbf{Full held-out batches for six occupation prompts.}
  Each panel shows all $M{=}16$ generated images for SD3.5-M (Base) and Ours,
  without selection.}
  \label{fig:app-full-batch-heldout}
\end{figure*}

Figure~\ref{fig:app-full-batch-heldout} shows the full $M{=}16$ batches of the race5 models for six held-out prompts, including \emph{underwriter}, without selection. These batches come from the race5 models, so the Ours labels for \emph{underwriter} differ from the race5$\times$gender2 batch in Figure~\ref{fig:race5-qual-batch}. FairFace labels
most Base images as White. For instance, the Base/Ours counts over White/Black/Asian/
Indian/Latino change from 11/0/1/1/3 to 1/1/4/3/7 for \emph{structural engineer},
from 13/0/1/0/2 to 4/3/4/2/3 for \emph{ecologist}, and from 11/0/0/1/4 to
2/3/5/2/4 for \emph{underwriter}. The Ours batches instead spread across all
five labels.

\clearpage
\section{Prompt-Level Finite-Batch Coverage}
\label{sec:app-finite-visibility}

Section~\ref{sec:relation} distinguishes pooled composition from finite-batch
coverage. Table~\ref{tab:finite-visibility} reports a hard-label version on the same $M{=}16$ held-out generations. \emph{Distinct@16} is the number of distinct FairFace labels in a prompt's 16-image batch, averaged over prompts. \emph{FullCov@16}
is the fraction of prompts whose batch contains every target mode or cell.
\emph{min Occ@16} is the minimum, over modes or cells, of the fraction of prompts where that mode or cell appears. FairFace is not used in training.

\begin{table}[t]
  \centering
\caption{\textbf{Prompt-level finite-batch coverage.}
Distinct@16, FullCov@16, and min Occ@16 are computed from FairFace hard top
labels on the same held-out $M{=}16$ batches used for
Table~\ref{tab:main-fairness}. RL rows report mean $\pm$ standard deviation across three runs.}
\label{tab:finite-visibility}
\scriptsize
\setlength{\tabcolsep}{2.2pt}
\resizebox{\ifdim\width>\linewidth\linewidth\else\width\fi}{!}{%
\begin{tabular}{@{}llcccc@{}}
\toprule
Labels & Method & Distinct@16 $\uparrow$ & FullCov@16 $\uparrow$ &
min Occ@16 $\uparrow$ & FairFace $F$ $\uparrow$\\
\midrule
race5 & SD3.5-M (Base)
& 3.32 & 9.0\% & 33.0\% & 0.494\\
& GRPO ($k{=}1$)~\citep{liu2025flowgrpo}
& $3.52{\pm}0.03$ & $15.5{\pm}1.5$\% & $46.7{\pm}0.2$\% &
$0.521{\pm}0.007$\\
& GRPO (count)~\citep{chen2026holofair}
& $3.58{\pm}0.08$ & $16.7{\pm}2.9$\% & $47.7{\pm}3.4$\% &
$0.561{\pm}0.013$\\
& \textbf{Ours}
& $3.70{\pm}0.02$ & $19.2{\pm}0.5$\% &
$47.3{\pm}2.8$\% & $0.771{\pm}0.013$\\
\midrule
race5$\times$gender2 & SD3.5-M (Base)
& 4.48 & 0.0\% & 13.5\% & 0.530\\
& GRPO ($k{=}1$)~\citep{liu2025flowgrpo}
& $4.73{\pm}0.02$ & $0.0{\pm}0.0$\% & $11.7{\pm}1.2$\% &
$0.557{\pm}0.011$\\
& GRPO (count)~\citep{chen2026holofair}
& $4.69{\pm}0.04$ & $0.2{\pm}0.2$\% & $16.8{\pm}1.7$\% &
$0.592{\pm}0.016$\\
& \textbf{Ours}
& $4.62{\pm}0.05$ & $0.2{\pm}0.2$\% &
$23.8{\pm}2.3$\% & $0.746{\pm}0.009$\\
\bottomrule
\end{tabular}
}
\end{table}

For race5, Ours has the highest Distinct@16 and FullCov@16, while its min Occ@16 is similar to GRPO (count) ($47.3\%$ vs.\ $47.7\%$). A uniform generator would reach a FullCov@16 of $86\%$ on race5 and $7\%$ on race5$\times$gender2.
For the ten race5$\times$gender2 cells, Distinct@16 is similar across methods and full coverage remains rare at $M{=}16$; the main gain is in min Occ@16, which rises to $23.8\%$ from $13.5\%$ for Base and $16.8\%$ for GRPO (count).

\section{Per-Mode Frequency Breakdown}
\label{sec:app-composition}

Table~\ref{tab:race5-fairface-frequency} gives the FairFace per-mode frequencies behind Figure~\ref{fig:race5-distribution-bars}, for the same run of each method.
Frequencies use images with a detected face; the no-face column gives the fraction of the 3,200 images per method with no detected face.

\begin{table}[t]
  \centering
  \caption{\textbf{FairFace per-mode frequency on race5.} Frequencies are over images with a detected face, as for the FairFace Fairness Score; no-face is the fraction of images without a detected face.}
  \label{tab:race5-fairface-frequency}
  \scriptsize
  \setlength{\tabcolsep}{4pt}
  \resizebox{\ifdim\width>\linewidth\linewidth\else\width\fi}{!}{
  \begin{tabular}{lcccccc}
    \toprule
    Method & White & Black & Asian & Indian & Latino & no-face\\
    \midrule
    SD3.5-M (Base) & 0.566 & 0.034 & 0.104 & 0.057 & 0.238 & 0.044\\
    FairImagen~\citep{fu2025fairimagen} & 0.537 & 0.038 & 0.114 & 0.061 & 0.250 & 0.047\\
    Weak Guidance~\citep{kim2025weak} & 0.321 & 0.098 & 0.095 & 0.197 & 0.289 & 0.052\\
    GRPO ($k{=}1$)~\citep{liu2025flowgrpo} & 0.587 & 0.057 & 0.113 & 0.053 & 0.190 & 0.107\\
    GRPO (count)~\citep{chen2026holofair} & 0.477 & 0.068 & 0.116 & 0.079 & 0.261 & 0.048\\
    \textbf{Ours} & 0.272 & 0.084 & 0.191 & 0.143 & 0.310 & 0.014\\
    \bottomrule
  \end{tabular}
  }
\end{table}

\section{Non-Person Prompts}
\label{sec:app-quality}

\begin{figure*}[t]
  \centering
  \includegraphics[width=\textwidth]{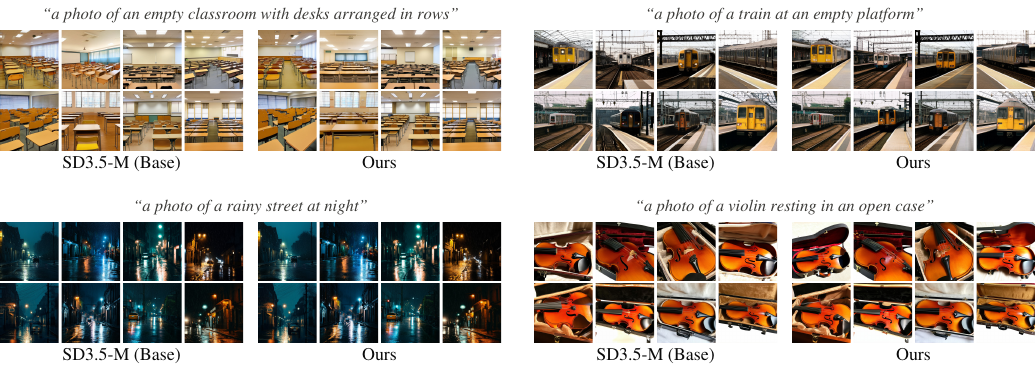}
  \caption{\textbf{Examples for non-person prompts.} Each panel shows
  SD3.5-M (Base) and Ours for the same prompt and seeds.}
  \label{fig:ood-nonperson-qual}
\end{figure*}

We check whether the race5 MultiMaxPO model adds people to images for prompts that do not mention people. We use 80 such prompts covering objects, empty interiors, landscapes, animals, food, instruments, and technical scenes, and generate eight images per prompt. For each image, Qwen2.5-VL answers yes or no to ``Is there a visible human person in this image?'' We report two rates: the fraction of images with a visible person (person images) and the fraction of prompts with at least one such image among the eight (any-person prompts). Ours does not add more people than the base model (1.6\% vs.\ 2.2\% of images, 3.8\% vs.\ 6.3\% of prompts), and PickScore, aesthetic score, and CLIP-T stay at the base level (Table~\ref{tab:ood-nonperson}). Figure~\ref{fig:ood-nonperson-qual} shows examples for the same prompts and seeds.

\begin{table}[t]
  \centering
  \caption{\textbf{Non-person prompts.} 80 prompts with eight images each. Visible-person rates are judged by the Qwen2.5-VL yes/no question in the text.}
  \label{tab:ood-nonperson}
  \scriptsize
  \setlength{\tabcolsep}{4pt}
  \resizebox{\ifdim\width>\linewidth\linewidth\else\width\fi}{!}{%
  \begin{tabular}{lccccc}
    \toprule
    Method & Person images (\%) & Any-person prompts (\%)
      & PickScore & Aesthetic & CLIP-T\\
    \midrule
    SD3.5-M (Base) & 2.2 & 6.3 & 0.870 & 5.54 & 0.274\\
    GRPO (count)~\citep{chen2026holofair}     & 1.6 & 6.3 & 0.870 & 5.54 & 0.275\\
    Ours     & 1.6 & 3.8 & 0.870 & 5.57 & 0.275\\
    \bottomrule
  \end{tabular}
  }
\end{table}

\section{Second Backbone: FLUX.1-dev}
\label{sec:app-flux}

We repeat the SD3.5-M experiments on FLUX.1-dev~\citep{blackforestlabs2024flux}.
We keep the training prompts, CLIP reward, set size $k{=}D$, KL
weight $\beta{=}0.05$, learning rate, 360 training steps, three training runs,
held-out prompts, evaluators, and quality metrics. The following settings differ.
FLUX.1-dev is guidance-distilled, so we use guidance 3.5 with 6 denoising steps
during training and 28 at evaluation. The LoRA has rank 64 and alpha 128 and
covers the attention and feed-forward layers of both the image and text
streams, and the PPO clip range is $10^{-4}$. Each optimizer step uses 32
images (two prompt groups) instead of 256, so over 360 steps the model sees
one-eighth as many prompt groups as SD3.5-M (720 vs.\ 5{,}760). Intervals below
are 95\% bootstrap intervals over the 200 held-out prompts, computed on the
Fairness Score averaged over the three runs.

\begin{figure*}[t]
  \centering
  \includegraphics[width=0.6\textwidth]{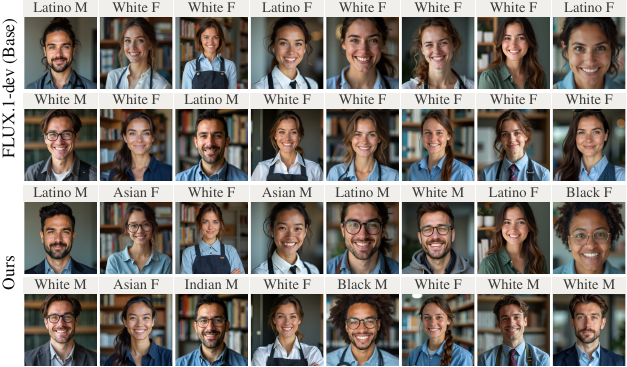}
  \caption{\textbf{FLUX.1-dev on race5$\times$gender2: full batches for \emph{bookkeeper}.}
  All 16 held-out images of FLUX.1-dev (Base) and Ours with the same seeds, with
  automatic FairFace labels. Base: 12 White and 4 Latino; 3 male and 13 female. Ours: 7/2/3/1/3 over White/Black/Asian/Indian/Latino; 9 male and 7 female, covering 9 of the
  10 joint cells.}
  \label{fig:flux-qual}
\end{figure*}

\paragraph{Race and race-by-gender.}
Table~\ref{tab:flux-race} shows the results. In both settings, Ours scores higher than the base model and GRPO (count) on all three evaluators, and image quality stays
about the same. All twelve intervals for Ours minus base and Ours minus GRPO (count)
are above zero; for race5 FairFace, the gain is $+0.194$ $[0.170, 0.218]$ over the base model and $+0.107$ $[0.095, 0.123]$ over GRPO (count). The gains are smaller
than on SD3.5-M, where each step uses eight times as many images, and race5
varies more across runs (FairFace $0.581{\pm}0.076$). GRPO ($k{=}1$) scores
below the base model in both settings. Figure~\ref{fig:flux-qual} shows the full
batches for one held-out prompt.

\begin{table}[t]
  \centering
  \caption{\textbf{FLUX.1-dev, race5 and race5$\times$gender2.}
  Same training prompts, reward, held-out prompts, and evaluators as
  Table~\ref{tab:main-fairness}. RL rows report mean $\pm$ standard deviation across three runs. Pick., Aes., and C-T are as in Table~\ref{tab:main-fairness}.}
  \label{tab:flux-race}
  \scriptsize
  \setlength{\tabcolsep}{2.2pt}
  \resizebox{\ifdim\width>\linewidth\linewidth\else\width\fi}{!}{%
  \begin{tabular}{@{}llcccccc@{}}
    \toprule
    Labels & Method & CLIP $F$ & VLM $F$ & FairFace $F$ & Pick. & Aes. & C-T\\
    \midrule
    race5 & FLUX.1-dev (Base) & 0.331 & 0.117 & 0.387 & 0.805 & 5.94 & 0.216\\
    & GRPO ($k{=}1$) & $\meanstd{0.299}{0.010}$ & $\meanstd{0.082}{0.012}$ & $\meanstd{0.364}{0.030}$ & 0.806 & 5.99 & 0.218\\
    & GRPO (count) & $\meanstd{0.397}{0.022}$ & $\meanstd{0.198}{0.020}$ & $\meanstd{0.474}{0.011}$ & 0.805 & 5.94 & 0.216\\
    & \textbf{Ours} & $\bmeanstd{0.473}{0.062}$ & $\bmeanstd{0.287}{0.079}$ & $\bmeanstd{0.581}{0.076}$ & 0.805 & 5.91 & 0.213\\
    \midrule
    race5$\times$gender2 & FLUX.1-dev (Base) & 0.405 & 0.247 & 0.416 & 0.805 & 5.94 & 0.216\\
    & GRPO ($k{=}1$) & $\meanstd{0.317}{0.090}$ & $\meanstd{0.234}{0.108}$ & $\meanstd{0.356}{0.077}$ & 0.795 & 5.91 & 0.211\\
    & GRPO (count) & $\meanstd{0.442}{0.060}$ & $\meanstd{0.327}{0.068}$ & $\meanstd{0.473}{0.050}$ & 0.804 & 5.96 & 0.217\\
    & \textbf{Ours} & $\bmeanstd{0.472}{0.042}$ & $\bmeanstd{0.371}{0.031}$ & $\bmeanstd{0.523}{0.041}$ & 0.803 & 5.91 & 0.217\\
    \bottomrule
  \end{tabular}
  }
\end{table}

\paragraph{Gender: where the gain does not transfer.}
Trained on the neutral-person prompts, Ours does not improve gender balance on
the held-out occupation prompts (Table~\ref{tab:flux-gender}; FairFace $0.701$
vs.\ $0.705$ for the base model). The two prompt sets are skewed in opposite
directions: on 50 sampled training prompts, base FLUX.1-dev generates 91\%
women (FairFace), while on the held-out occupation prompts it generates 65\%
men. On the training prompts themselves, the share of prompts whose 16 images include both genders rises from 48\% to 62\% (one run).
However, this push toward more men does not help on the male-skewed occupation
prompts: Ours leaves the male share at 65\%, and GRPO (count) raises it to 71\%.

When we instead train on 1{,}000 occupation prompts (200 occupations, none of
them in the held-out set), Ours improves FairFace from $0.705$ to
$0.795{\pm}0.009$. GRPO (count) also improves,
to $0.762{\pm}0.056$. Ours is higher than GRPO (count) under FairFace ($+0.033$
$[0.010, 0.055]$), but the intervals include zero under CLIP and VLM. GRPO (count) produces more prompts with both genders (61\% vs.\ 56\%), while Ours gives the
more balanced pooled share. We report these rows separately because they use
different training prompts from the rest of the paper.

The failure on neutral prompts is not an evaluator error: the CLIP gender label agrees with FairFace on about 98\% of FLUX.1-dev images and with VLM on over 99\%. The gain carries over to the held-out prompts when the training prompts are
skewed in the same direction.

\begin{table}[t]
  \centering
  \caption{\textbf{FLUX.1-dev, binary gender, by training prompts.}
  All rows are evaluated on the same 200 held-out occupation prompts. Men and
  both genders are FairFace shares: the fraction of images labeled male, and
  the fraction of prompts whose 16 images include both genders (means over
  three runs). GRPO ($k{=}1$) is omitted because it receives zero advantage on gender2 (Table~\ref{tab:gender2-control}).}
  \label{tab:flux-gender}
  \scriptsize
  \setlength{\tabcolsep}{2.2pt}
  \resizebox{\ifdim\width>\linewidth\linewidth\else\width\fi}{!}{%
  \begin{tabular}{@{}llccccc@{}}
    \toprule
    Training prompts & Method & CLIP $F$ & VLM $F$ & FairFace $F$ & Men & Both genders\\
    \midrule
    -- & FLUX.1-dev (Base) & 0.741 & 0.738 & 0.705 & 65\% & 52\%\\
    \midrule
    neutral-person & GRPO (count) & $\meanstd{0.641}{0.043}$ & $\meanstd{0.639}{0.045}$ & $\meanstd{0.591}{0.044}$ & 71\% & 57\%\\
    & Ours & $\meanstd{0.732}{0.022}$ & $\meanstd{0.731}{0.022}$ & $\meanstd{0.701}{0.018}$ & 65\% & 52\%\\
    \midrule
    occupation & GRPO (count) & $\meanstd{0.810}{0.047}$ & $\meanstd{0.806}{0.047}$ & $\meanstd{0.762}{0.056}$ & 62\% & 61\%\\
    & \textbf{Ours} & $\bmeanstd{0.825}{0.008}$ & $\bmeanstd{0.823}{0.007}$ & $\bmeanstd{0.795}{0.009}$ & 60\% & 56\%\\
    \bottomrule
  \end{tabular}
  }
\end{table}

\section{Computational Cost}
\label{sec:app-compute}
All SD3.5-M fine-tuning runs in our main experiments train only a LoRA adapter. The reported checkpoint at step 360 was trained on 32 NVIDIA H100 GPUs and took about 19--23 GPU-hours in total per run. GRPO ($k{=}1$), GRPO (count), and Ours use the same number of rollouts and policy updates and differ only in how advantages are computed from the per-axis scores $r_d(x_i,c)$. Unless otherwise stated, all evaluations use the same inference-time sampling budget as the base model.

This is a one-time training cost. At inference, our method needs no reranking, best-of-$N$ selection, or per-image edits to prompts or embeddings: the LoRA weights can be merged into the base model, so sampling is unchanged.

\end{document}